\documentclass[11pt]{article}
\usepackage[final]{acl}

\usepackage{times}
\usepackage{latexsym}
\usepackage[T1]{fontenc}
\usepackage[utf8]{inputenc}
\usepackage{microtype}
\usepackage{inconsolata}
\usepackage{booktabs}
\usepackage{amsmath}
\usepackage{amsfonts}
\usepackage{nicefrac}
\usepackage{multirow}
\usepackage{xcolor}
\usepackage{colortbl}
\usepackage{graphicx}
\usepackage{listings}
\usepackage{tcolorbox}
\usepackage{bm}
\usepackage{pifont}
\usepackage{xspace}
\usepackage{url}

\graphicspath{{images/}}

\definecolor{desc}{RGB}{99,178,238}
\definecolor{acc}{RGB}{118,218,145}
\definecolor{rej}{RGB}{248,149,136}
\definecolor{uclablue}{rgb}{0.15,0.45,0.68}
\definecolor{c0}{RGB}{0,120,212}

\newcommand{\method}{\textsl{LogitSpec}\xspace}
\newcommand{\blue}{\cellcolor{uclablue!15}}

\newcommand{\greenyes}{\textcolor{green}{\ding{51}}}
\newcommand{\redno}{\textcolor{red}{\ding{55}}}
\newcommand{\good}{\textcolor{acc}}

\newtcbox{\mybox}[1][yellow]{on line, arc = 0pt, outer arc = 0pt,
  colback = #1!10!white, colframe = #1!50!black,
  boxsep = 0pt, left = 1pt, right = 1pt, top = 1pt, bottom = 1pt,
  boxrule = 0pt, bottomrule = 1pt, toprule = 1pt, fontupper = \ttfamily}

\hypersetup{
  breaklinks,
  colorlinks=true,
  linkcolor=uclablue,
  citecolor=uclablue
}

\title{\method: Accelerating Retrieval-based Speculative Decoding via Next Next Token Speculation}

\author{
   Tianyu Liu$^{1,2}$\thanks{Equal Contribution.} \quad Qitan Lv$^{1,2\textcolor{uclablue}{\ast}}$ \quad Hao Li$^{1,2}$ \quad Xing Gao$^2$ \quad Xiao Sun$^{2}$\thanks{The Corresponding Author.} \quad Xiaoyan Sun$^{1}$ \quad 
  \\
  ${}^1$University of Science and Technology of China \quad
  ${}^2$Shanghai AI Laboratory \\
    \texttt{\{tianyu\_liu, qitanlv, lihaohn\}@mail.ustc.edu.cn} \\
    \texttt{sunxiao@pjlab.org.cn}
    \quad \texttt{sunxiaoyan@ustc.edu.cn}
}

\begin{document}
\maketitle

\begin{abstract}
Speculative decoding (SD), where a small draft model is employed to propose \textit{draft} tokens in advance and then the target model validates them in parallel, has emerged as a promising technique for LLM inference acceleration.
Many endeavors to improve SD are to eliminate the need for a draft model and generate draft tokens in a retrieval-based manner in order to further alleviate the drafting overhead and significantly reduce the difficulty in deployment and applications.
However, retrieval-based SD relies on a matching paradigm to retrieve the most relevant reference as the draft tokens, where these methods often fail to find matched and accurate draft tokens.
To address this challenge, we propose \method to effectively expand the retrieval range and find the most relevant reference as drafts. \method is motivated by the observation that the logit of the last token can not only predict \textbf{the next token}, but also speculate \textbf{the next next token}.
Specifically, \method generates draft tokens in two steps: (1) utilizing the last logit to speculate the next next token; (2) retrieving relevant reference for both the next token and the next next token.
\method is training-free and plug-and-play, which can be easily integrated into existing LLM inference frameworks.
Extensive experiments on a wide range of text generation benchmarks demonstrate that \method can achieve up to 2.61$\times$ speedup and 3.28 mean accepted tokens per decoding step. Our code is available at \url{https://github.com/smart-lty/LogitSpec}.

\end{abstract}

\section{Introduction}\label{intro}
Large Language Models (LLMs), such as GPT-4.5 \citep{gpt4}, DeepSeek R1 \citep{r1}, Qwen 2.5 \citep{qwen2.5}, and LLaMA-3 \citep{llama3}, have demonstrated remarkable capabilities across a wide range of natural language processing tasks, including question answering \citep{qa}, code generation \citep{code_generation}, and dialogue systems \citep{dialogue_system}.
While they achieve success by scaling up the model parameters, the increase in scale comes with significant inference challenges.
The most straightforward challenge is \textit{auto-regressive decoding}, where each LLM decoding step can only generate one token.
This token-by-token decoding process incurs exacerbated latency with both the length of generated tokens and the model scale.

To address this challenge, speculative decoding (SD) \citep{sps1, sps2} has emerged as a promising approach for lossless LLM inference acceleration.
The key idea of SD is to employ a small draft model to first generate $\gamma$ draft tokens, and then the target model validates these tokens in parallel within a single forward, ensuring the output distribution to be unchanged.
However, deploying an extra draft model introduces considerable overhead and difficulties, including (a) \underline{\textbf{complex implementation and deployment}}, especially in integration with other techniques,\citep{pearl} (b) \underline{\textbf{additional memory overhead}}, especially in long-context scenarios. Moreover, when there exists no available draft model for SD, it takes \underline{\textbf{substantial training cost}} to construct a compact draft model \citep{eagle3}. (Please refer to Section \ref{related_work} for more details.)

Therefore, many existing works are dedicated to developing draft-model-free SD methods \citep{pld, lade, rest}, where the draft tokens are generated in a retrieval-based manner. At each decoding step, they retrieve the relevant reference  \textbf{according to the last few tokens} and extract draft tokens from the reference. In this way, they can explicitly eliminate the need for a draft model and reduce the drafting overhead.

\begin{figure*}[t]
    \centering
    \includegraphics[width=\textwidth]{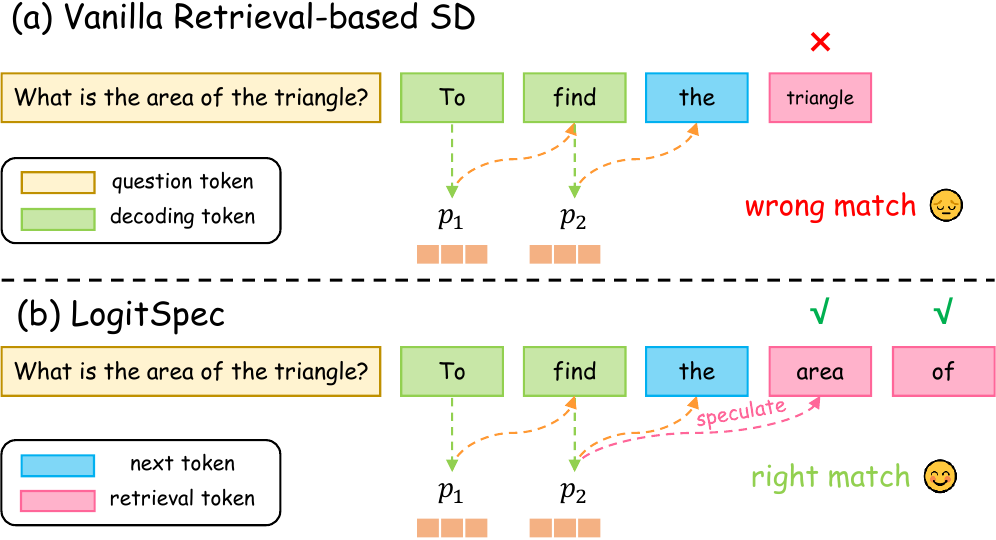}
    \caption{Illustration of vanilla retrieval-based SD method and our \method. (a) Retrieval-based SD retrieves the wrong token ``\textbf{\textcolor{red}{triangle}}'' according to the next token ``the''. (b) \method first speculates the next next token ``area'', and then retrieves the right relevant reference ``\textbf{\good{area of}}'' according to ``the area''. This simple example illustrates how \method utilizes the last logit to speculate the next next token and improves the retrieval accuracy.}
    \label{fig:last_logit_spec}
\end{figure*}
Despite their promising efficacy and implementation simplicity, these methods rely on a matching paradigm to effectively retrieve the most relevant reference as draft tokens and often struggle to find matched and accurate tokens. As shown in Figure \ref{fig:last_logit_spec} (a), with a simple prompt ``\textit{What is the area of the triangle?}'', vanilla retrieval-based SD methods fail to effectively retrieve the right token ``area'' according to the next token ``the'', as the most relevant reference is ``the triangle''. Moreover, it is often the case that no matched reference can be found, e.g., there are no matched tokens in more than 30\% of decoding steps in PLD \citep{pld}.

In this paper, we propose a simple yet effective retrieval-based SD framework, namely \method, which leverages the logit of the last token (\textit{last logit}) to predict the next next token and improves the accuracy of the retrieval reference. \method is motivated by the observation that \textbf{the last logit can speculate the next next token with a relatively high accuracy.} Based on this observation, we use the speculated next next token as guidance to retrieve reference. Specifically, \method generates draft tokens in two steps: (1) utilizing the last logit to speculate the next next token; (2) retrieving relevant reference for both the next token and the next next token. As shown in Figure \ref{fig:last_logit_spec} (b), with the speculated token ``area'', \method successfully retrieves the correct draft tokens ``area of''. This two-step process enables better retrieval accuracy---it can help filter the most relevant reference when the reference is redundant, while extending the searching space when there exists no relevant reference. To summarize, our contributions are:

\begin{itemize}
    \item [(a)] We empirically observe that the last logit can speculate the next next token with a relatively high accuracy. Unlike other retrieval techniques, which are sensitive to the specific task, this property is \textbf{robust} and \textbf{effective} across various tasks.
    \item [(b)] Building on this observation, we propose \method, a \textbf{plug-and-play} retrieval-based SD framework which can improve the retrieval accuracy and achieve better speedup.
    \item [(c)] We conduct extensive experiments on various text generation benchmarks to demonstrate the effectiveness of \method. Notably, \method achieves up to 2.61$\times$ speedup and 3.28 mean accepted tokens per decoding step without the need for an extra draft model.
\end{itemize}

\section{Related Work}\label{related_work}
In this section, we briefly review the two main directions of speculative decoding, draft-model-based speculative decoding and draft-model-free speculative decoding.
More discussions are provided in Appendix \ref{app:related_work}.

\paragraph{Draft-model-based speculative decoding.}
This line of research focuses on improving the draft quality, which primarily invests heavily in post-training a specialized draft model to generate high-quality drafts.
Medusa \citep{medusa} pioneers this direction by adding extra MLP heads at the top of the target model, reusing the hidden states from the target model, and predicting the next few tokens in parallel. However, the generation from these decoding heads is independent, which harms the accuracy of the draft tokens. Therefore, some works \citep{hydra,clover} propose to add sequential dependencies for better performance. Glide \citep{glide} takes a similar idea of Medusa and proposes to reuse the KV cache from the target model. Recently, Eagle \citep{eagle, eagle2, eagle3} has dominated this research line by training an auto-regressive head and scaling up the training data. \textbf{However, while these methods achieve superior speedup, all of them rely on an extra draft model, which necessitates extra parameters or extensive training.} The existence of the draft model significantly limits its application: (a) deploying a draft model requires complex and careful implementation. (b) Both the draft model and its KV cache require additional GPU memory, which may incur load imbalance. In the long context settings, even the KV cache for the draft model may exceed the capacity of 1 GPU and lead to significant resource competitions; (c) the training cost is substantial, and the draft model requires retraining when the target model is updated.

\paragraph{Draft-model-free speculative decoding.}
{This line of research focuses on maximizing drafting efficiency by eliminating the need for a draft model entirely, aiming for universal, zero-cost acceleration.}
Some existing works \citep{draft_and_verify, layerskip, swift} observe that the layer sparsity of LLMs means that not all the layers are necessary to predict the next token. They propose methods to generate draft tokens with a subset of the layers of the target model. However, the number of skipped layers is limited, which affects the overall speedup. Recently, retrieval-based SD \citep{rest, pld, lade} has dominated the draft-model-free speculative decoding by retrieving the relevant reference as the draft tokens. The relevant reference can be derived from the question and generated tokens, or an external database. \textbf{However, there exist two common issues during the retrieval process}: (a) retrieval from the context often fails to find matched draft tokens; (b) retrieval from an external database often struggles to find accurate draft tokens. These two issues motivate us to develop a more appropriate retrieval-based speculative decoding framework. To the best of our knowledge, \method is the first to use a speculated next-next token as an additional retrieval anchor; this mechanism is training-free and complementary to multi-token prediction methods that require trained extra heads.

\section{Background}\label{background}

\paragraph{Speculative Decoding.} Let $\bm{x}$ denote an input sequence (prefix). A speculative decoding step consists of a drafting phase and a verification phase. During the drafting phase, the draft model $\mathcal{M}_q$ is employed to give $\gamma$ draft tokens $x_1, x_2,...,x_\gamma$ by running $\gamma$ times the model forward and sampling. Here, we denote the output logit $\mathcal{M}_q(\bm{x}+[x_1,...,x_{i-1}])$ as $q_{i-1}$, then each draft token is given by
$x_i \sim q_{i-1}, i=1,...,\gamma$. During the verification phase, the prefix $\bm{x}$ together with $\gamma$ draft tokens are sent to $\mathcal{M}_p$ for verification. The target model $\mathcal{M}_p$ inputs $\bm{x}+[x_1, ..., x_\gamma]$ and outputs the logits $p_0, p_1, ..., p_{\gamma}$. Then SD sequentially verifies $x_i$ via speculative sampling \citep{sps1,sps2}, where the acceptance rate is given by:
\begin{equation}
    \alpha_i=\left\{
    \begin{aligned}
        &1 \quad\qquad\quad p_{i-1}[x_i] \geq q_{i-1}[x_i], \\
        &\frac{p_{i-1}[x_i]}{q_{i-1}[x_i]} \quad p_{i-1}[x_i] < q_{i-1}[x_i],\\
    \end{aligned}
    \right.
\end{equation}

If SD rejects $x_i$, it will resample a token from $norm(\max(0, p_{i-1} - q_{i-1}))$, otherwise, SD accepts all the draft tokens and samples an additional token from $p_{\gamma}$. In this way, each SD step generates at least 1 token and at most $\gamma+1$, leading to theoretical lossless quality and efficiency acceleration.

\paragraph{Retrieval-based Speculative Decoding.} The retrieval-based SD methods generate draft tokens in a retrieval-based manner. As retrieval-based SD methods do not require an additional draft model, we omit the notation for $\mathcal{M}_q$ and abbreviate $\mathcal{M}_p$ as $\mathcal{M}$. A retrieval model $\mathcal{R}$ is employed to store $n$-grams (referred as \textbf{reference}) from corpus
\begin{equation}
    \mathcal{R} = \{(x_i^{1}, x_i^{2}, ..., x_i^{n})\}_{i=1}^{N}.
\end{equation}

Here, $N$ denotes the total number of $n$-grams in $\mathcal{R}$. Existing retrieval-based SD methods mainly differ in constructing $\mathcal{R}$. At each SD decoding step, given a query $m$-gram tokens $x_q^1, x_q^2, ..., x_q^m$, the retrieval model first traverses the reference, finds matched $n$-grams and returns the subsequent tokens of the matched tokens:

\begin{multline}
    \textsc{Match}\big(\mathcal{R},(x_q^1, x_q^2, ..., x_q^m)\big) \\
    = \{(x_i^{m+1}, ..., x_i^{n}) \mid i\in\mathcal{S}\}.
\end{multline}

where $\mathcal{S}=\{i|x_i^t=x_q^t, \forall t=1,...,m\}$. Intuitively, a larger $m$ leads to more precise matches but lower match probabilities.

\section{Method}
\begin{figure}[t]
\centering
\includegraphics[width=\columnwidth]{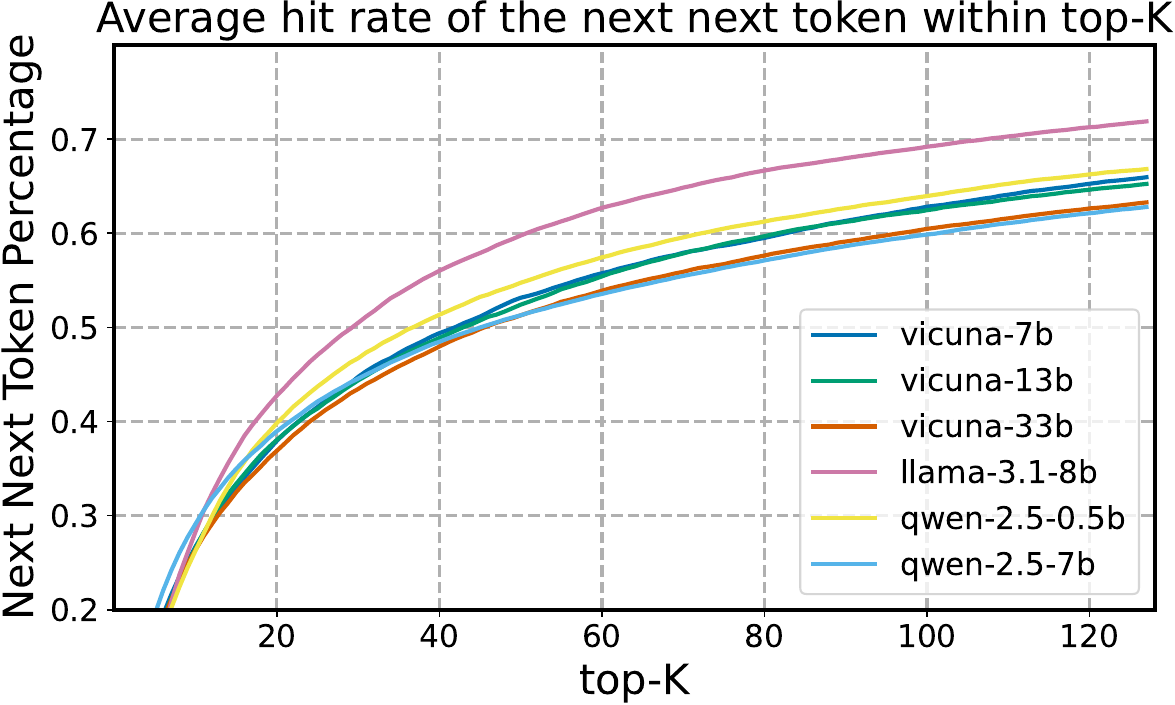}
\caption{Motivated observation I: the last logit can speculate the next next token with high accuracy. In over 50\% decoding steps, the next next token can be found in the top-60 entries within the last logit across model sizes and architectures.}
\label{fig:motivation_acc}
\end{figure}

\begin{figure*}[t]
    \centering
    \includegraphics[width=1.0\textwidth]{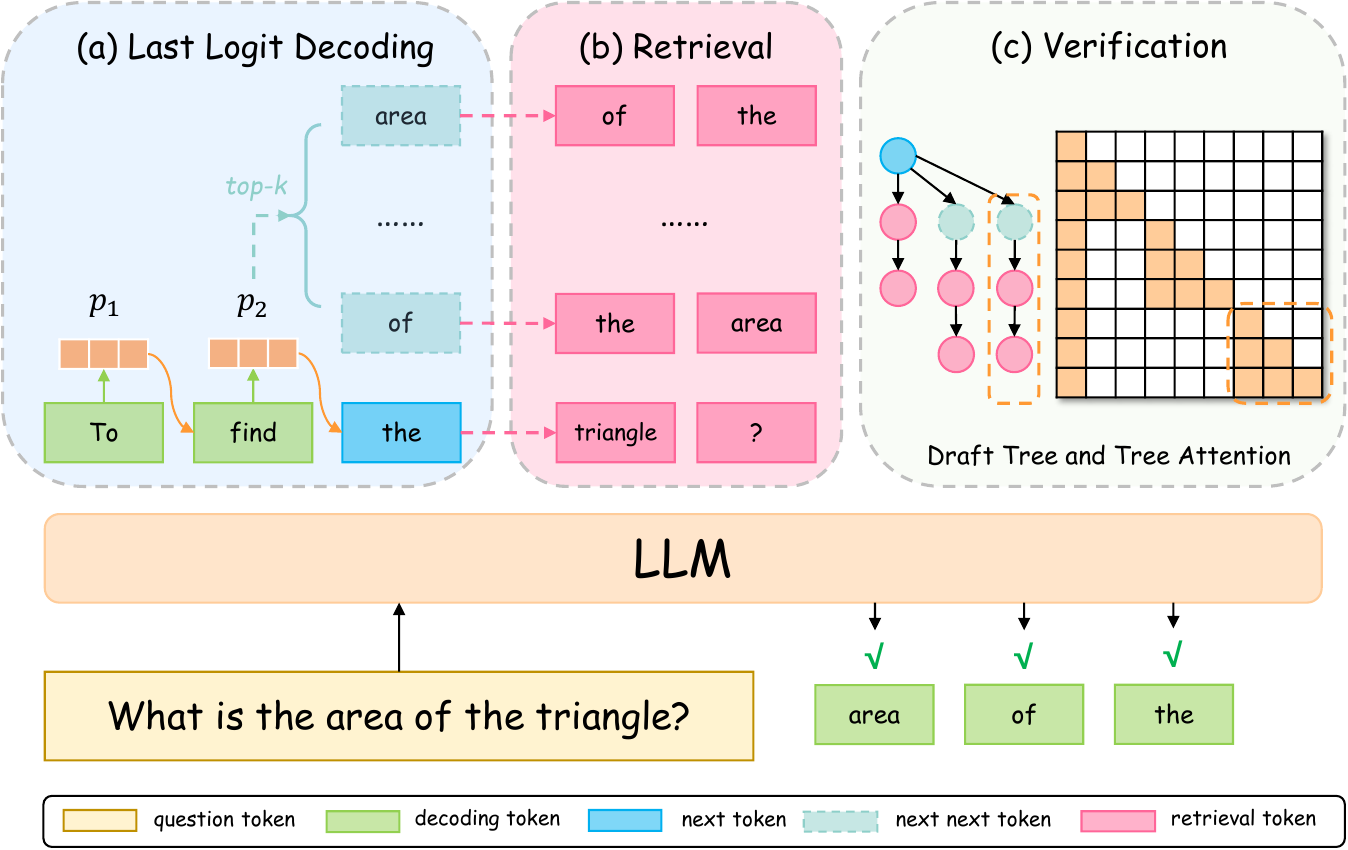}
    \caption{An overview of \method. At each decoding step, \method first utilizes the top-$k$ entries of the last logit as the speculation for the next next token. Then, \method retrieves relevant references for both the next token and the next next token. Finally, \method organizes the draft tokens into a draft tree and prepares a tree attention for parallel verification.}
    \label{fig:model_fig}
\end{figure*}

In this section, we introduce our \method, a novel retrieval-based SD framework that utilizes \textit{last logit} as a guidance for retrieval and effectively improves the retrieval performance. We first conduct a motivated experiment in Section \ref{motivated_exp}, and then introduce the whole framework of \method in Section \ref{draft_sec} and \ref{verify_sec}. An overview of \method is shown in Figure \ref{fig:model_fig}.
\subsection{Motivated Observation}\label{motivated_exp}
Drawing insights from multi-token prediction \citep{mtp} that fine-tuned LLMs can predict multiple tokens in a single forward, we are interested in the ability of LLMs to predict the next next token \textbf{without fine-tuning}. Unlike MTP, which operates at the hidden-state level and requires training extra heads, we study the logit-level signal of off-the-shelf LLMs (see Appendix~\ref{app:motivation_exp} for an intuition). We conduct a simple experiment to investigate the rank of the next next token in the last logit, which is used to predict the next token. Specifically, we use a small fraction of Spec-Bench \citep{spec-bench} that contains 13 sub-tasks (detailed in Appendix \ref{app:motivation_exp}) and 6 different LLMs as backbones. Results in Figure \ref{fig:motivation_acc} demonstrate that \textbf{the last logit has a strong potential to predict the next next token}. In over 50\% decoding steps, the next next token can be found within the top-60 entries of the last logit. For Llama 3.1 8B \citep{llama3} with a large 128K vocabulary, this percentage even increases to 64\% within the top-60 entries of the last logit. The results of 6 different LLMs on 13 downstream tasks demonstrate that this observation holds consistently across different models (Vicuna, Llama 3.1, and Qwen) and scales (ranging from 0.5B to 33B).

\begin{figure}[t]
\centering
\includegraphics[width=\columnwidth]{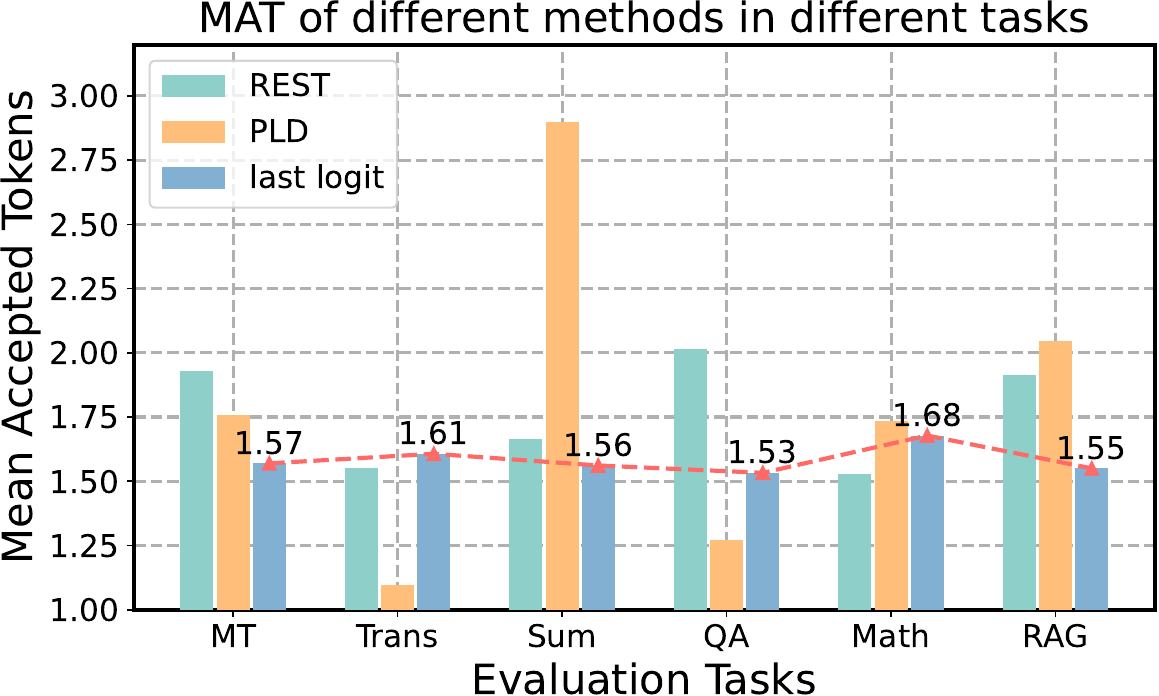}
\caption{Motivated observation II: compared with other retrieval-based methods, the prediction of the last logit demonstrates robustness to downstream tasks, motivating us to utilize it to guide the retrieval process.}
\label{fig:motivation_mat}
\end{figure}

We further conduct experiments on Spec-Bench with Vicuna 7B to investigate the robustness of the predictive capability of the last logit across different tasks. We propose a minimal speculative decoding algorithm with last logit, \textit{last logit decoding} for this experiment, where we directly use the top-60 entries of the last logit to serve as the draft tokens. Each draft token is a guess at the next next token. If the target model accepts any one of the draft tokens, the others are dropped. Consequently, at each decoding step, the accepted length is either 2 or 1. As in Figure \ref{fig:motivation_mat}, this simple SD method yields mean accepted tokens per decoding step (MAT) of over 1.5. Compared with other retrieval-based SD methods such as PLD \citep{pld} and REST \citep{rest}, which are highly affected by the task type, \textbf{the MAT of \textit{last logit decoding} exhibits superior robustness to the task.} These results demonstrate that the ability of the last logit to predict the next next token widely exists in modern LLMs themselves, and strongly motivate us to improve retrieval-based SD methods with the last logit. We provide a detailed experimental setup for the motivation experiments in Appendix \ref{app:motivation_exp}.

\subsection{\method Drafting}\label{draft_sec}
While \textit{last logit decoding} has shown great simplicity in implementation and robustness to the downstream tasks, its theoretical upper bound of the speedup ratio is 2, as all the draft tokens are guesses for the next next token. In our experiments, the \textit{last logit decoding} achieves an overall speedup ratio of $1.2\times\sim1.4\times$,
which hinders the real-world applications of \textit{last logit decoding} due to the unsatisfactory speedup ratio.
Therefore, to further improve the overall inference speedup, we propose to utilize the prediction ability of the last logit as \textbf{a guidance for retrieval}. Specifically, for a given input prefix $\mathbf{x}=(x_1, x_2, ..., x_i)$ and the last logit $p_i=\mathcal{M}(x_1, x_2, ..., x_i)$, we first sample the next token $x_{i+1} \sim p_i$. Then, we utilize the top-$k$ entries of the last logit (except the next token $x_{i+1}$) as the speculation to the next next token $\widetilde{x}_{i+2}$:
\begin{equation}
    \mathcal{L}
    = \{\widetilde{x}_{i+2} \mid \widetilde{x}_{i+2} \in \textsc{Top}_k(p_i),\, \widetilde{x}_{i+2} \neq x_{i+1}\}.
\end{equation}

Then, we use the retrieval model $\mathcal{R}$ to retrieve reference for both the next token and next next tokens:

\begin{align}
    & \mathcal{D}_{i+1}(x_{i+1}) = \textsc{Match}\big(\mathcal{R}, \bm{g}^{(1)}_i(x_{i+1})\big), && \label{eq:match_next}\\
    & \mathcal{D}_{i+2}(\widetilde{x}_{i+2}) = \textsc{Match}\big(\mathcal{R}, \bm{g}^{(2)}_i(\widetilde{x}_{i+2})\big), && \label{eq:match_nextnext}\\
    & \mathcal{D} = \mathcal{D}_{i+1}(x_{i+1}) \cup \bigcup_{\widetilde{x}_{i+2}\in \mathcal{L}} \mathcal{D}_{i+2}(\widetilde{x}_{i+2}). && \label{eq:match_union}
\end{align}

Here, $\bm{g}^{(1)}_i(x_{i+1})=(x_{i-m+2}, ..., x_{i}, x_{i+1})$ denotes the $m$-gram query for the next token, and $\bm{g}^{(2)}_i(\widetilde{x}_{i+2})=(x_{i-m+3}, ..., x_{i}, x_{i+1}, \widetilde{x}_{i+2})$ denotes the query for the next next token. In our experiments, we retrieve draft tokens with $m=3$ first. If no matched tokens are found, we decrease $m$ to 2 and so on. For simplicity, we only use the user-input prompt and decoded tokens to construct the retrieval model $\mathcal{R}$, that is to say, \textbf{our \method is query-independent, and will not be affected by other queries.} We give the detailed implementation of the retrieval process in Appendix \ref{app:retrieval}.

Taking the example in Figure \ref{fig:model_fig}, we first select the top-$k$ entries of the last logit as the speculation for the next next token. Then, we retrieve reference for both the next token ``the'' and the next next token ``area'', and return ``of the''. After that, we organize these draft tokens into a draft tree, where the next token serves as the root. Finally, the token sequence ``area of the'' is accepted by $\mathcal{M}$.

After the retrieval process, we employ a simple pruning strategy to control the number of draft tokens. Specifically, we do not prune the retrieval tokens for the next token. For each speculated next next token, we prune the retrieval tokens with a simple heuristic strategy: if the rank of the next next token is below 8, we preserve 4 tokens; if the rank of the next next token is below 32, we preserve 3 tokens; otherwise, we only preserve the speculated next next token itself. The retrieval process is terminated until the total number of draft tokens exceeds a specific draft tree capacity ${K}$. These thresholds are chosen as simple fractions of $K$ without careful tuning, implementing a pyramid allocation that gives more retrieved tokens to higher-confidence speculations. Furthermore, we explored the impact of various pruning strategies, with results detailed in Table~\ref{tab:llama3_pruning_ab} of Appendix~\ref{more_pruning}.

\subsection{\method Verification}\label{verify_sec}

After retrieving multiple draft token sequences $\mathcal{D}$, we organize these token sequences into a draft tree and prepare a tree attention for parallel verification.

Then, we prepare an attention mask to make each draft token sequence invisible to other sequences. Let $\Lambda(l)$ denote a causal mask with length $l$, and $m_j$ denote the length of the $j$-th token sequence in $\mathcal{D}$; the attention mask is given by:

\begin{equation}
    A_{\text{draft}} = \operatorname{diag}\big(\Lambda(m_1), \ldots, \Lambda(m_j)\big).
\end{equation}
An illustration of this attention mask is shown in Figure \ref{fig:model_fig}(c) (note that the next token is visible to all sequences, corresponding to the first column). Both the preparation of the attention mask and the verification of the draft tree are consistent with previous works \citep{specinfer, medusa}. We provide a pseudo code of the attention mask preparations in Appendix \ref{app:pseudocode}.

\section{Experiments}

\subsection{Experimental Setup}\label{sec:exp_set}

\paragraph{Tasks and Datasets.}
We evaluate \method on a broad suite of text generation benchmarks. We first use Spec-Bench \citep{spec-bench}, a widely used comprehensive benchmark that covers diverse application scenarios, including multi-turn conversation (MT), translation (Trans), summarization (Sum), question answering (QA), mathematical reasoning (Math), and retrieval-augmented generation (RAG). We then further evaluate \method on HumanEval \citep{humaneval}, GSM8K \citep{gsm8k}, CNN/DM \citep{cnndm}, MATH \citep{math}, AIME 24\&25 \citep{AIME}, and LongBench \citep{longbench}, which are widely used benchmarks for code generation, math reasoning, summarization, and long-context generation.

\paragraph{Implementation Details.} We follow standard inference settings with temperature 0 and batch size 1. All implementation details (baselines, hardware, software versions, model configurations) are provided in Appendix \ref{app:implementation}. We use two widely used metrics for evaluation: mean accepted tokens per decoding step (denoted as MAT) and overall speedup ratio (denoted as Speedup). We further evaluate \method under non-zero temperatures ($T>0$) in Appendix~\ref{app:sampling}.

\subsection{Main Results}

We conduct experiments on various text generation benchmarks to demonstrate the effectiveness of \method. As in Table \ref{tab:vicuna_3_bench} (with the Spec-Bench breakdown deferred to Appendix \ref{app:vicuna_specbench}), \method outperforms retrieval-based SD baselines by a large margin. We adopt the Vicuna series as the main-table backbone because existing baselines (e.g., REST, Lookahead) only provide stable, reproducible implementations for the Vicuna/Llama-2 family; results on modern Llama-3.1 and Qwen3 backbones are reported in Appendix~\ref{more_backbone}.
We further present more in-depth analysis of \method including (i) more LLM backbones (Llama2 series, Llama-3.1-Instruct-8B and Qwen-3-8B), (ii) more benchmarks (MATH and AIME datasets), and (iii) long context benchmarks (the LongBench dataset) in Appendix \ref{app:more_exp_results}. Comparisons against draft-model-based methods (EAGLE, Medusa) and stronger retrieval-based baselines (Token Recycling, SAM Decoding) are deferred to Appendix~\ref{app:draft_model_based} and Appendix~\ref{more_backbone}, respectively.

\begin{table*}[!t]
\centering
\caption{Experimental results of \method on CNN/DM \citep{cnndm}, GSM8K \citep{gsm8k} and HumanEval \citep{humaneval} with \textbf{Vicuna}. We report the mean accepted tokens per decoding step (MAT) and overall speedup ratio. We \textbf{bold} the best results and \underline{underline} the suboptimal results for each backbone model. \\}
\label{tab:vicuna_3_bench}
\resizebox{\linewidth}{!}{
\begin{tabular}{lccccccccc|cc}
\toprule
\multirow{2}{*}{\textbf{Models}} & \multirow{2}{*}{\textbf{Method}} & \multicolumn{2}{c}{\textbf{CNN/DM}} & \multicolumn{2}{c}{\textbf{GSM8K}} & \multicolumn{2}{c}{\textbf{HumanEval}} & \multicolumn{2}{c}{\textbf{Spec-Bench}} & \multicolumn{2}{c}{\textbf{Average}} \\
\cmidrule(lr){3-4} \cmidrule(lr){5-6} \cmidrule(lr){7-8} \cmidrule(lr){9-10} \cmidrule(lr){11-12}
& & MAT  & Speedup & MAT  & Speedup & MAT  & Speedup & MAT & Speedup & MAT  & Speedup\\

\midrule
\multirow{5}{*}{\textbf{Vicuna 7B}}
    & Lookahead  & 1.54&1.28&1.92&\underline{1.64}&1.79&1.57 & 1.66 & 1.27 & 1.73 & 1.44 \\
    & REST   & 1.64&1.19&1.55&1.13&1.96&1.51 & 1.82 & 1.48 & 1.74 & 1.33\\
    & PLD  & \underline{2.61}&\underline{2.26}&1.80&1.62&1.84&1.65 & 1.73 & \underline{1.59} & 2.00 & \underline{1.78} \\
    & SpS & 2.34&1.58 &\underline{2.08}&1.49&\underline{2.56}&\underline{1.80} & \underline{2.28} & 1.49 & \underline{2.32} & 1.59\\
    & \blue{\textbf{\method}} & \blue{\textbf{3.28}} & \blue{\textbf{2.61}} & \blue{\textbf{2.69}} & \blue{\textbf{2.20}} & \blue{\textbf{2.62}} & \blue{\textbf{2.24}} & \blue{\textbf{2.44}} & \blue{\textbf{2.01}} & \blue{\textbf{2.76}} & \blue{\textbf{2.26}}\\

\midrule
\multirow{5}{*}{\textbf{Vicuna 13B}}
    & Lookahead  & 1.46&1.12&1.88&\underline{1.61}&1.75&1.57&1.63&1.22 & 1.68&1.38\\
    & REST   & 1.65&1.22&1.57&1.18&1.94&1.55&1.82&1.38 & 1.75&1.33\\
    & PLD  & \underline{2.34}&\underline{1.85}&1.76&1.61&1.98&1.77&1.68&1.48 & 1.94&1.68 \\
    & SpS & 2.18&1.48&\underline{2.00}&1.55&\underline{2.66}&\underline{1.95}&\underline{2.18}&\underline{1.49} & \underline{2.26}&1.62 \\
    & \blue{\textbf{\method}} & \blue{\textbf{2.90}} & \blue{\textbf{2.17}} & \blue{\textbf{2.59}} & \blue{\textbf{2.13}} & \blue{\textbf{2.88}} & \blue{\textbf{2.47}} & \blue{\textbf{2.32}} & \blue{\textbf{1.93}} & \blue{\textbf{2.67}} & \blue{\textbf{2.18}} \\

\midrule
\multirow{5}{*}{\textbf{Vicuna 33B}}
    & Lookahead  & 1.50&1.18&1.89&1.44&1.66&1.33&1.61&1.24 & 1.67&1.30\\
    & REST   &1.65&1.41&1.57&1.32&1.99&1.64&1.80&1.48 & 1.75&1.46\\
    & PLD  & \underline{2.07}&\underline{1.72}&1.66&1.41&1.60&1.42&1.55&1.42 & 1.72&1.49 \\
    & SpS & 2.01&1.57&\underline{1.97}&\underline{1.54}&\underline{2.27}&\underline{1.80}&\underline{2.01}&\underline{1.61} & \underline{2.07}&\underline{1.63} \\
& \blue{\textbf{\method}} & \blue{\textbf{2.66}} & \blue{\textbf{2.01}} & \blue{\textbf{2.46}} & \blue{\textbf{1.93}} & \blue{\textbf{2.41}} & \blue{\textbf{1.95}} & \blue{\textbf{2.13}} & \blue{\textbf{1.76}} & \blue{\textbf{2.42}} & \blue{\textbf{1.92}}\\

\bottomrule
\end{tabular}
}
\end{table*}

We observe the following: (a) \method significantly improves MAT to $2.13\sim 3.28$, achieving up to $2.61\times$ speedup on CNN/DM with Vicuna 7B and $2.47\times$ on HumanEval with Vicuna 13B, substantially outperforming REST (<1.5$\times$ on CNN/DM). (b) Thanks to the robustness of last logit speculation, \method maintains strong performance even in challenging scenarios with limited context repetition, where other retrieval-based methods struggle (see detailed per-task analysis in Appendix \ref{app:vicuna_specbench}). (c) \method significantly outperforms PLD across all benchmarks and model sizes, demonstrating that next-next token speculation effectively guides retrieval and improves draft quality, especially when output has low overlap with input context.

\subsection{Ablation Study}

To further provide more insights of \method, we conduct extensive ablation studies on the aforementioned 4 benchmarks in Table \ref{tab:vicuna_ablation}. Specifically, we mainly focus on two components of \method: \textit{last logit} and \textit{retrieval}. We denote \method without the last logit decoding and only using retrieval as \textit{w/o last logit}, and \method without retrieval and only using last logit decoding as \textit{w/o retrieval}. As shown in Table \ref{tab:vicuna_ablation}, the absence of any component results in a performance degradation of the entire framework.

Our findings are as follows. First, the absence of \textit{retrieval} exhibits more importance to the final acceleration, which is consistent with the discussion in Section \ref{draft_sec} that the theoretical upper bound of \textit{last logit decoding} severely hinders its real-world application. Second, the absence of \textit{retrieval} and the absence of \textit{last logit} show different effects in different sub-tasks. For example, the absence of \textit{retrieval} decreases MAT by 0.87 on Spec-Bench with Vicuna 7B, while it decreases MAT by 1.71 on CNN/DM. In contrast, the absence of \textit{last logit} leads to a more consistent MAT degradation across different tasks, which further highlights the robustness of \textit{last logit decoding}. Finally, these results demonstrate that combining \textit{last logit} with \textit{retrieval} improves the retrieval performance and overall speedup.
{We also conduct ablation studies on pruning strategies in Appendix~\ref{more_pruning} and~\ref{more_capacity}. The experimental results suggest that varying the pruning strategy yields only minor differences.}
\begin{table*}[!t]
\centering
\caption{Ablation experiments of \method on Spec-Bench \citep{spec-bench}, CNN/DM \citep{cnndm}, GSM8K \citep{gsm8k} and HumanEval \citep{humaneval} with \textbf{Vicuna}. We report the ablation results with MAT reduction and Speedup reduction with down arrow $\color{red}{\downarrow}$.\\}
\label{tab:vicuna_ablation}
\resizebox{\linewidth}{!}{
\begin{tabular}{llllllllll}
\toprule
\multirow{2}{*}{\textbf{Models}} & \multirow{2}{*}{\textbf{Method}} & \multicolumn{2}{c}{\textbf{Spec-Bench}} & \multicolumn{2}{c}{\textbf{CNN/DM}} & \multicolumn{2}{c}{\textbf{GSM8K}} & \multicolumn{2}{c}{\textbf{HumanEval}} \\
\cmidrule(lr){3-4} \cmidrule(lr){5-6} \cmidrule(lr){7-8} \cmidrule(lr){9-10}
& & MAT  & Speedup & MAT  & Speedup & MAT  & Speedup & MAT  & Speedup\\

\midrule
\multirow{3}{*}{\textbf{Vicuna 7B}}
    & \textit{w/o last logit} & 1.72$_{\color{red}{\downarrow.72}}$ & 1.27$_{\color{red}{\downarrow.74}}$ & 2.66$_{\color{red}{\downarrow.62}}$ & 2.24$_{\color{red}{\downarrow.37}}$ & 1.81$_{\color{red}{\downarrow.88}}$ & 1.63$_{\color{red}{\downarrow.57}}$ & 1.89$_{\color{red}{\downarrow.73}}$ & 1.69$_{\color{red}{\downarrow.55}}$ \\
    & \textit{w/o retrieval}  & 1.57$_{\color{red}{\downarrow.87}}$ & 1.24$_{\color{red}{\downarrow.77}}$ & 1.57$_{\color{red}{\downarrow1.71}}$ & 1.23$_{\color{red}{\downarrow1.38}}$ & 1.71$_{\color{red}{\downarrow.98}}$ & 1.39$_{\color{red}{\downarrow.81}}$ & 1.69$_{\color{red}{\downarrow.93}}$ & 1.39$_{\color{red}{\downarrow.85}}$ \\
    & \blue{\textbf{\method}} & \blue{\textbf{2.44}} & \blue{\textbf{2.01}} & \blue{\textbf{3.28}} & \blue{\textbf{2.61}} & \blue{\textbf{2.69}} & \blue{\textbf{2.20}} & \blue{\textbf{2.62}} & \blue{\textbf{2.24}} \\

\midrule
\multirow{3}{*}{\textbf{Vicuna 13B}}
    & \textit{w/o last logit} & 1.66$_{\color{red}{\downarrow.66}}$ & 1.23$_{\color{red}{\downarrow.70}}$ & 2.34$_{\color{red}{\downarrow.56}}$ & 1.81$_{\color{red}{\downarrow.36}}$ & 1.78$_{\color{red}{\downarrow.81}}$ & 1.61$_{\color{red}{\downarrow.52}}$ & 2.02$_{\color{red}{\downarrow.86}}$ & 1.75$_{\color{red}{\downarrow.72}}$ \\
    & \textit{w/o retrieval}  & 1.59$_{\color{red}{\downarrow.73}}$ & 1.19$_{\color{red}{\downarrow.74}}$ & 1.58$_{\color{red}{\downarrow1.32}}$ & 1.22$_{\color{red}{\downarrow.95}}$ & 1.71$_{\color{red}{\downarrow.88}}$ & 1.45$_{\color{red}{\downarrow.68}}$ & 1.63$_{\color{red}{\downarrow1.25}}$ & 1.37$_{\color{red}{\downarrow1.10}}$ \\
    & \blue{\textbf{\method}} & \blue{\textbf{2.32}} & \blue{\textbf{1.93}} & \blue{\textbf{2.90}} & \blue{\textbf{2.17}} & \blue{\textbf{2.59}} & \blue{\textbf{2.13}} & \blue{\textbf{2.88}} & \blue{\textbf{2.47}} \\

\midrule
\multirow{3}{*}{\textbf{Vicuna 33B}}
    & \textit{w/o last logit} & 1.55$_{\color{red}{\downarrow.58}}$ & 1.22$_{\color{red}{\downarrow.54}}$ & 2.05$_{\color{red}{\downarrow.61}}$ & 1.72$_{\color{red}{\downarrow.29}}$ & 1.64$_{\color{red}{\downarrow.82}}$ & 1.39$_{\color{red}{\downarrow.54}}$ & 1.59$_{\color{red}{\downarrow.82}}$ & 1.40$_{\color{red}{\downarrow.55}}$\\
    & \textit{w/o retrieval}  & 1.61$_{\color{red}{\downarrow.52}}$ & 1.25$_{\color{red}{\downarrow.51}}$ & 1.62$_{\color{red}{\downarrow1.04}}$ & 1.26$_{\color{red}{\downarrow.75}}$ & 1.69$_{\color{red}{\downarrow.77}}$ & 1.33$_{\color{red}{\downarrow.60}}$ & 1.60$_{\color{red}{\downarrow.81}}$ & 1.34$_{\color{red}{\downarrow.61}}$\\
    & \blue{\textbf{\method}} & \blue{\textbf{2.13}} & \blue{\textbf{1.76}} & \blue{\textbf{2.66}} & \blue{\textbf{2.01}} & \blue{\textbf{2.46}} & \blue{\textbf{1.93}} & \blue{\textbf{2.41}} & \blue{\textbf{1.95}}\\

\bottomrule
\end{tabular}
}
\end{table*}

\begin{table*}[htbp]
\centering

\caption{Case study experiments of \method on Spec-Bench and HumanEval with \textbf{Vicuna}. We report the successful retrieval rate of each method. We report the relative improvements with $\color{desc}{\uparrow}$.\\}
\label{tab:vicuna_retrieval_case_study}
\resizebox{\linewidth}{!}{
\begin{tabular}{lllllll}
\toprule
 \multirow{2}{*}{\textbf{Method}} & \multicolumn{3}{c}{\textbf{Spec-Bench}} & \multicolumn{3}{c}{\textbf{HumanEval}} \\
\cmidrule(lr){2-4} \cmidrule(lr){5-7}
& Vicuna 7B & Vicuna 13B & Vicuna 33B & Vicuna 7B & Vicuna 13B & Vicuna 33B \\

\midrule
PLD & 63.88 & 63.08 & 62.71 & 69.03 & 69.51 & 67.67 \\
\method & 97.76$_{\color{desc}{\uparrow53.04\%}}$ & 97.64$_{\color{desc}{\uparrow54.79\%}}$ & 97.93$_{\color{desc}{\uparrow56.16\%}}$ & 99.29$_{\color{desc}{\uparrow43.84\%}}$ & 99.31$_{\color{desc}{\uparrow42.87\%}}$ & 99.37$_{\color{desc}{\uparrow46.84\%}}$\\

\bottomrule
\end{tabular}
}
\end{table*}

\subsection{Case Study}

\begin{figure}[t]
    \centering
    \includegraphics[width=0.9\columnwidth]{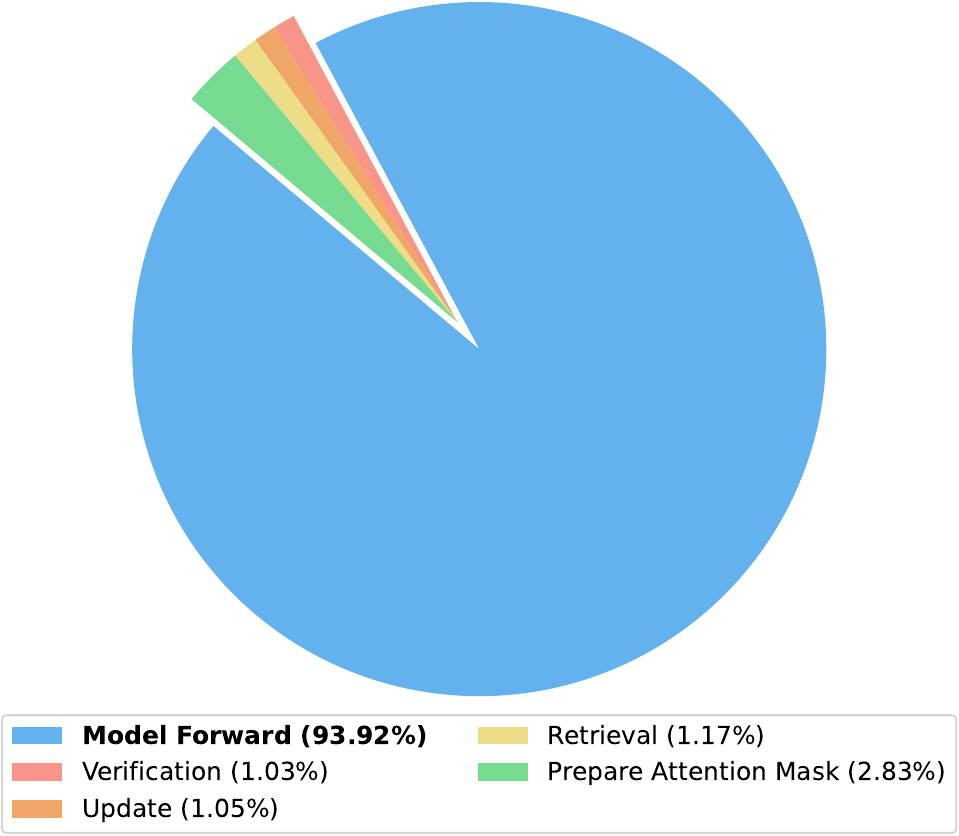}
    \caption{Running time breakdown of the whole decoding process on Spec-Bench with Vicuna 7B.}
    \label{fig:runtime}
\end{figure}

\paragraph{In-depth Running Time Analysis.}  We conduct experiments to analyze the running time allocation of the whole decoding. Specifically, there are \textbf{five non-negligible components} in \method, including (a) \textit{retrieving draft tokens}: the process of retrieving reference for the next token and the next next tokens; (b) \textit{preparation}: preparing attention mask for the draft tokens; (c) \textit{model forward}: conducting one-pass model forward; (d) \textit{verification}: validating the draft tokens with speculative sampling; (e) \textit{update}: necessary update of KV cache and retrieval model.

As shown in Figure \ref{fig:runtime}, \textit{model forward} occupies the majority of wall-clock time. Compared with vanilla AR decoding, the overhead introduced by \method, i.e., retrieving overhead, only takes 1.17\% of the whole decoding process. It can be further alleviated by parallel techniques, as the retrieval process is independent,
which brings negligible overhead as well.

\paragraph{Retrieval Performance.} As mentioned in Section \ref{draft_sec}, \method expands the retrieval range and improves the retrieval performance. We conduct a simple experiment to test the retrieval success rate, i.e., whether the model successfully retrieves the reference as draft tokens. As in Table \ref{tab:vicuna_retrieval_case_study}, PLD cannot retrieve any matched tokens in more than 30\% decoding steps, while \method retrieves matched reference in most decoding steps. These results further demonstrate the effectiveness of \method.

{
\paragraph{Real-world examples.} We also provide a real-world example in Appendix~\ref{more_realworld} to illustrate how \textit{next-next token speculation} allows \method to succeed where standard retrieval methods fail.
}

\section{Conclusion}\label{sec:conclusion}
In this paper, we  observe that the logit of the last token can predict \textbf{the next next token} with a relatively high accuracy without \textbf{any fine-tuning}.
Based upon this observation, we propose a novel retrieval-based SD framework, namely \method, which utilizes the prediction ability of the last logit to effectively expand the retrieval range and find the most relevant reference as the draft tokens. \method does not require an additional draft model and is a fully \textbf{{plug-and-play}} method, which can be easily implemented and integrated into existing LLM frameworks. Extensive experiments demonstrate that \method can effectively improve the retrieval performance, leading to a $1.8\times \sim 2.6\times$ speedup across all the evaluation benchmarks.

\section*{Limitations}
\label{sec:limitation}
We consider a few limitations and future works as follows. \textbf{(i)} While our \method is a fully plug-and-play SD framework, its real-world inference acceleration is less competitive. Our future works involve integrating \method into existing draft-model-based SD methods for further acceleration. \textbf{(ii)} Currently, \method retrieves relevant reference from the prompt, which may incur lower speedup when the prompt is short. Concretely, \method still achieves $\sim$$1.4\times$ speedup on translation tasks where context $n$-grams are scarce (Table~\ref{tab:vicuna_specbench}) and rises to $2.01\times$ on LongBench with longer contexts (Appendix~\ref{more_longdata}). We consider integrating an external database to boost the retrieval model as a future work. \textbf{(iii)} Our acceleration estimates rely on offline decoding with curated prompts and NVIDIA A100 GPUs, so the reported speedups may vary on other workloads or hardware. \method currently retrieves draft candidates from the on-the-fly prompt context only; short prompts or out-of-domain corpora reduce its benefits.

\section*{Ethical Considerations}
\label{sec:ethic}
\paragraph{Inheritance of Model Behaviors.} \method is an inference acceleration framework designed to speed up existing Large Language Models without modifying their weights. As a speculative decoding method, it aims to losslessly recover the distribution of the target model. Consequently, \method inherits the ethical properties, biases, and potential safety risks of the underlying target LLM and draft model. It does not introduce new capabilities for generating harmful content, nor does it mitigate existing biases in the base models. Users should continue to apply standard safety guardrails and alignment techniques to the target models deployed with \method.

\clearpage

\bibliography{logitspec_acl}

@InProceedings{sps1,
  title = 	 {Fast Inference from Transformers via Speculative Decoding},
  author =       {Leviathan, Yaniv and Kalman, Matan and Matias, Yossi},
  booktitle = 	 {Proceedings of the 40th International Conference on Machine Learning},
  pages = 	 {19274--19286},
  year = 	 {2023},
  editor = 	 {Krause, Andreas and Brunskill, Emma and Cho, Kyunghyun and Engelhardt, Barbara and Sabato, Sivan and Scarlett, Jonathan},
  volume = 	 {202},
  series = 	 {Proceedings of Machine Learning Research},
  month = 	 {23--29 Jul},
  publisher =    {PMLR},
  url = 	 {https://proceedings.mlr.press/v202/leviathan23a.html}
}

@misc{sps2,
      title={Accelerating Large Language Model Decoding with Speculative Sampling}, 
      author={Charlie Chen and Sebastian Borgeaud and Geoffrey Irving and Jean-Baptiste Lespiau and Laurent Sifre and John Jumper},
      year={2023},
      eprint={2302.01318},
      archivePrefix={arXiv},
      primaryClass={cs.CL},
      url={https://arxiv.org/abs/2302.01318}, 
}

@misc{gpt4,
      title={GPT-4 Technical Report}, 
      author={OpenAI},
      year={2024},
      eprint={2303.08774},
      archivePrefix={arXiv},
      primaryClass={cs.CL},
      url={https://arxiv.org/abs/2303.08774}, 
}

@misc{r1,
      title={DeepSeek-R1: Incentivizing Reasoning Capability in LLMs via Reinforcement Learning}, 
      author={DeepSeek-AI},
      year={2025},
      eprint={2501.12948},
      archivePrefix={arXiv},
      primaryClass={cs.CL},
      url={https://arxiv.org/abs/2501.12948}, 
}

@misc{llama3,
      title={The Llama 3 Herd of Models}, 
      author={Llama Team},
      year={2024},
      eprint={2407.21783},
      archivePrefix={arXiv},
      primaryClass={cs.AI},
      url={https://arxiv.org/abs/2407.21783}, 
}

@misc{qwen2.5,
      title={Qwen2.5 Technical Report}, 
      author={Qwen Team},
      year={2025},
      eprint={2412.15115},
      archivePrefix={arXiv},
      primaryClass={cs.CL},
      url={https://arxiv.org/abs/2412.15115}, 
}

@article{qa,
title = {A literature review on question answering techniques, paradigms and systems},
journal = {Journal of King Saud University - Computer and Information Sciences},
volume = {32},
number = {6},
pages = {635-646},
year = {2020},
issn = {1319-1578},
doi = {https://doi.org/10.1016/j.jksuci.2018.08.005},
url = {https://www.sciencedirect.com/science/article/pii/S131915781830082X},
author = {Marco Antonio {Calijorne Soares} and Fernando Silva Parreiras}
}

@misc{code_generation,
      title={A Survey on Large Language Models for Code Generation}, 
      author={Juyong Jiang and Fan Wang and Jiasi Shen and Sungju Kim and Sunghun Kim},
      year={2024},
      eprint={2406.00515},
      archivePrefix={arXiv},
      primaryClass={cs.CL},
      url={https://arxiv.org/abs/2406.00515}, 
}

@misc{dialogue_system,
      title={A Survey on Recent Advances in LLM-Based Multi-turn Dialogue Systems}, 
      author={Zihao Yi and Jiarui Ouyang and Yuwen Liu and Tianhao Liao and Zhe Xu and Ying Shen},
      year={2024},
      eprint={2402.18013},
      archivePrefix={arXiv},
      primaryClass={cs.CL},
      url={https://arxiv.org/abs/2402.18013}, 
}

@misc{eagle,
      title={EAGLE: Speculative Sampling Requires Rethinking Feature Uncertainty}, 
      author={Yuhui Li and Fangyun Wei and Chao Zhang and Hongyang Zhang},
      year={2025},
      eprint={2401.15077},
      archivePrefix={arXiv},
      primaryClass={cs.LG},
      url={https://arxiv.org/abs/2401.15077}, 
}

@inproceedings{eagle2,
    title = "{EAGLE}-2: Faster Inference of Language Models with Dynamic Draft Trees",
    author = "Li, Yuhui  and
      Wei, Fangyun  and
      Zhang, Chao  and
      Zhang, Hongyang",
    editor = "Al-Onaizan, Yaser  and
      Bansal, Mohit  and
      Chen, Yun-Nung",
    booktitle = "Proceedings of the 2024 Conference on Empirical Methods in Natural Language Processing",
    month = nov,
    year = "2024",
    address = "Miami, Florida, USA",
    publisher = "Association for Computational Linguistics",
    url = "https://aclanthology.org/2024.emnlp-main.422/",
    doi = "10.18653/v1/2024.emnlp-main.422",
    pages = "7421--7432"
}

@misc{eagle3,
      title={EAGLE-3: Scaling up Inference Acceleration of Large Language Models via Training-Time Test}, 
      author={Yuhui Li and Fangyun Wei and Chao Zhang and Hongyang Zhang},
      year={2025},
      eprint={2503.01840},
      archivePrefix={arXiv},
      primaryClass={cs.CL},
      url={https://arxiv.org/abs/2503.01840}, 
}

@inproceedings{medusa,
author = {Cai, Tianle and Li, Yuhong and Geng, Zhengyang and Peng, Hongwu and Lee, Jason D. and Chen, Deming and Dao, Tri},
title = {MEDUSA: Simple LLM inference acceleration framework with multiple decoding heads},
year = {2024},
publisher = {JMLR.org},
booktitle = {Proceedings of the 41st International Conference on Machine Learning},
articleno = {203},
numpages = {27},
location = {Vienna, Austria},
series = {ICML'24}
}

@inproceedings{
hydra,
title={Hydra: Sequentially-Dependent Draft Heads for Medusa Decoding},
author={Zachary Ankner and Rishab Parthasarathy and Aniruddha Nrusimha and Christopher Rinard and Jonathan Ragan-Kelley and William Brandon},
booktitle={First Conference on Language Modeling},
year={2024},
url={https://openreview.net/forum?id=FbhjirzvJG}
}

@misc{clover,
      title={Clover: Regressive Lightweight Speculative Decoding with Sequential Knowledge}, 
      author={Bin Xiao and Chunan Shi and Xiaonan Nie and Fan Yang and Xiangwei Deng and Lei Su and Weipeng Chen and Bin Cui},
      year={2024},
      eprint={2405.00263},
      archivePrefix={arXiv},
      primaryClass={cs.CL},
      url={https://arxiv.org/abs/2405.00263}, 
}

@inproceedings{specinfer,
author = {Miao, Xupeng and Oliaro, Gabriele and Zhang, Zhihao and Cheng, Xinhao and Wang, Zeyu and Zhang, Zhengxin and Wong, Rae Ying Yee and Zhu, Alan and Yang, Lijie and Shi, Xiaoxiang and Shi, Chunan and Chen, Zhuoming and Arfeen, Daiyaan and Abhyankar, Reyna and Jia, Zhihao},
title = {SpecInfer: Accelerating Large Language Model Serving with Tree-based Speculative Inference and Verification},
year = {2024},
isbn = {9798400703867},
publisher = {Association for Computing Machinery},
address = {New York, NY, USA},
url = {https://doi.org/10.1145/3620666.3651335},
doi = {10.1145/3620666.3651335},
booktitle = {Proceedings of the 29th ACM International Conference on Architectural Support for Programming Languages and Operating Systems, Volume 3},
pages = {932--949},
numpages = {18},
location = {La Jolla, CA, USA},
series = {ASPLOS '24}
}

@inproceedings{
hass,
title={Learning Harmonized Representations for Speculative Sampling},
author={Lefan Zhang and Xiaodan Wang and Yanhua Huang and Ruiwen Xu},
booktitle={The Thirteenth International Conference on Learning Representations},
year={2025},
url={https://openreview.net/forum?id=T9u56s7mbk}
}

@misc{judge,
      title={Judge Decoding: Faster Speculative Sampling Requires Going Beyond Model Alignment}, 
      author={Gregor Bachmann and Sotiris Anagnostidis and Albert Pumarola and Markos Georgopoulos and Artsiom Sanakoyeu and Yuming Du and Edgar Schönfeld and Ali Thabet and Jonas Kohler},
      year={2025},
      eprint={2501.19309},
      archivePrefix={arXiv},
      primaryClass={cs.LG},
      url={https://arxiv.org/abs/2501.19309}, 
}

@misc{coral,
      title={CORAL: Learning Consistent Representations across Multi-step Training with Lighter Speculative Drafter}, 
      author={Yepeng Weng and Dianwen Mei and Huishi Qiu and Xujie Chen and Li Liu and Jiang Tian and Zhongchao Shi},
      year={2025},
      eprint={2502.16880},
      archivePrefix={arXiv},
      primaryClass={cs.CL},
      url={https://arxiv.org/abs/2502.16880}, 
}

@misc{gumiho,
      title={Gumiho: A Hybrid Architecture to Prioritize Early Tokens in Speculative Decoding}, 
      author={Jinze Li and Yixing Xu and Haiduo Huang and Xuanwu Yin and Dong Li and Edith C. H. Ngai and Emad Barsoum},
      year={2025},
      eprint={2503.10135},
      archivePrefix={arXiv},
      primaryClass={cs.CL},
      url={https://arxiv.org/abs/2503.10135}, 
}

@misc{
specdec++,
title={SpecDec++: Boosting Speculative Decoding via Adaptive Candidate Lengths},
author={Kaixuan Huang and Xudong Guo and Mengdi Wang},
year={2025},
url={https://openreview.net/forum?id=NnExMNiTHw}
}

@misc{opt-tree,
      title={OPT-Tree: Speculative Decoding with Adaptive Draft Tree Structure}, 
      author={Jikai Wang and Yi Su and Juntao Li and Qingrong Xia and Zi Ye and Xinyu Duan and Zhefeng Wang and Min Zhang},
      year={2024},
      eprint={2406.17276},
      archivePrefix={arXiv},
      primaryClass={cs.CL},
      url={https://arxiv.org/abs/2406.17276}, 
}

@misc{adaeagle,
      title={AdaEAGLE: Optimizing Speculative Decoding via Explicit Modeling of Adaptive Draft Structures}, 
      author={Situo Zhang and Hankun Wang and Da Ma and Zichen Zhu and Lu Chen and Kunyao Lan and Kai Yu},
      year={2024},
      eprint={2412.18910},
      archivePrefix={arXiv},
      primaryClass={cs.AI},
      url={https://arxiv.org/abs/2412.18910}, 
}

@misc{speed,
      title={SPEED: Speculative Pipelined Execution for Efficient Decoding}, 
      author={Coleman Hooper and Sehoon Kim and Hiva Mohammadzadeh and Hasan Genc and Kurt Keutzer and Amir Gholami and Sophia Shao},
      year={2024},
      eprint={2310.12072},
      archivePrefix={arXiv},
      primaryClass={cs.CL},
      url={https://arxiv.org/abs/2310.12072}, 
}

@misc{free,
      title={Fast and Robust Early-Exiting Framework for Autoregressive Language Models with Synchronized Parallel Decoding}, 
      author={Sangmin Bae and Jongwoo Ko and Hwanjun Song and Se-Young Yun},
      year={2023},
      eprint={2310.05424},
      archivePrefix={arXiv},
      primaryClass={cs.CL},
      url={https://arxiv.org/abs/2310.05424}, 
}

@misc{eesd,
      title={Speculative Decoding via Early-exiting for Faster LLM Inference with Thompson Sampling Control Mechanism}, 
      author={Jiahao Liu and Qifan Wang and Jingang Wang and Xunliang Cai},
      year={2024},
      eprint={2406.03853},
      archivePrefix={arXiv},
      primaryClass={cs.CL},
      url={https://arxiv.org/abs/2406.03853}, 
}

@misc{tokenrecycling,
      title={Turning Trash into Treasure: Accelerating Inference of Large Language Models with Token Recycling}, 
      author={Xianzhen Luo and Yixuan Wang and Qingfu Zhu and Zhiming Zhang and Xuanyu Zhang and Qing Yang and Dongliang Xu and Wanxiang Che},
      year={2024},
      eprint={2408.08696},
      archivePrefix={arXiv},
      primaryClass={cs.CL},
      url={https://arxiv.org/abs/2408.08696}, 
}

@misc{samdecoding,
      title={SAM Decoding: Speculative Decoding via Suffix Automaton}, 
      author={Yuxuan Hu and Ke Wang and Xiaokang Zhang and Fanjin Zhang and Cuiping Li and Hong Chen and Jing Zhang},
      year={2024},
      eprint={2411.10666},
      archivePrefix={arXiv},
      primaryClass={cs.CL},
      url={https://arxiv.org/abs/2411.10666}, 
}

@inproceedings{glide,
author = {Du, Cunxiao and Jiang, Jing and Yuanchen, Xu and Wu, Jiawei and Yu, Sicheng and Li, Yongqi and Li, Shenggui and Xu, Kai and Nie, Liqiang and Tu, Zhaopeng and You, Yang},
title = {GLIDE with a CAPE: a low-hassle method to accelerate speculative decoding},
year = {2024},
publisher = {JMLR.org},
booktitle = {Proceedings of the 41st International Conference on Machine Learning},
articleno = {465},
numpages = {17},
location = {Vienna, Austria},
series = {ICML'24}
}

@misc{layerskip,
    title={LayerSkip: Enabling Early Exit Inference and Self-Speculative Decoding},
    author={Mostafa Elhoushi and Akshat Shrivastava and Diana Liskovich and Basil Hosmer and Bram Wasti and Liangzhen Lai and Anas Mahmoud and Bilge Acun and Saurabh Agarwal and Ahmed Roman and Ahmed A Aly and Beidi Chen and Carole-Jean Wu},
    booktitle = "Proceedings of the 62nd Annual Meeting of the Association for Computational Linguistics (Volume 1: Long Papers)",
    month = aug,
    year = "2024",
    address = "Bangkok, Thailand",
    publisher = "Association for Computational Linguistics",
    url = "https://aclanthology.org/2024.acl-long.681",
    doi = "10.18653/v1/2024.acl-long.681",
    pages = "12622--12642",
}

@inproceedings{draft_and_verify,
    title = "Draft {\&} Verify: Lossless Large Language Model Acceleration via Self-Speculative Decoding",
    author = "Zhang, Jun  and
      Wang, Jue  and
      Li, Huan  and
      Shou, Lidan  and
      Chen, Ke  and
      Chen, Gang  and
      Mehrotra, Sharad",
    editor = "Ku, Lun-Wei  and
      Martins, Andre  and
      Srikumar, Vivek",
    booktitle = "Proceedings of the 62nd Annual Meeting of the Association for Computational Linguistics (Volume 1: Long Papers)",
    month = aug,
    year = "2024",
    address = "Bangkok, Thailand",
    publisher = "Association for Computational Linguistics",
    url = "https://aclanthology.org/2024.acl-long.607/",
    doi = "10.18653/v1/2024.acl-long.607",
    pages = "11263--11282"
}

@inproceedings{rest,
    title = "{REST}: Retrieval-Based Speculative Decoding",
    author = "He, Zhenyu  and
      Zhong, Zexuan  and
      Cai, Tianle  and
      Lee, Jason  and
      He, Di",
    editor = "Duh, Kevin  and
      Gomez, Helena  and
      Bethard, Steven",
    booktitle = "Proceedings of the 2024 Conference of the North American Chapter of the Association for Computational Linguistics: Human Language Technologies (Volume 1: Long Papers)",
    month = jun,
    year = "2024",
    address = "Mexico City, Mexico",
    publisher = "Association for Computational Linguistics",
    url = "https://aclanthology.org/2024.naacl-long.88/",
    doi = "10.18653/v1/2024.naacl-long.88",
    pages = "1582--1595"
}

@inproceedings{lade,
author = {Fu, Yichao and Bailis, Peter and Stoica, Ion and Zhang, Hao},
title = {Break the sequential dependency of LLM inference using LOOKAHEAD DECODING},
year = {2024},
publisher = {JMLR.org},
booktitle = {Proceedings of the 41st International Conference on Machine Learning},
articleno = {561},
numpages = {20},
location = {Vienna, Austria},
series = {ICML'24}
}

@misc{pld,
    title = {Prompt Lookup Decoding},
    author = {Apoorv Saxena},
    year = {2023},
    month = {November},
    url = {https://github.com/apoorvumang/prompt-lookup-decoding/}
}

@misc{mtp,
      title={Better \& Faster Large Language Models via Multi-token Prediction}, 
      author={Fabian Gloeckle and Badr Youbi Idrissi and Baptiste Rozière and David Lopez-Paz and Gabriel Synnaeve},
      year={2024},
      eprint={2404.19737},
      archivePrefix={arXiv},
      primaryClass={cs.CL},
      url={https://arxiv.org/abs/2404.19737}, 
}

@inproceedings{
pearl,
title={{PEARL}: Parallel Speculative Decoding with Adaptive Draft Length},
author={Tianyu Liu and Yun Li and Qitan Lv and Kai Liu and Jianchen Zhu and Winston Hu and Xiao Sun},
booktitle={The Thirteenth International Conference on Learning Representations},
year={2025},
url={https://openreview.net/forum?id=QOXrVMiHGK}
}

@inproceedings{spec-bench,
    title = "Unlocking Efficiency in Large Language Model Inference: A Comprehensive Survey of Speculative Decoding",
    author = "Xia, Heming and Yang, Zhe and Dong, Qingxiu and Wang, Peiyi and Li, Yongqi  and Ge, Tao and Liu, Tianyu and Li, Wenjie and Sui, Zhifang",
    editor = "Ku, Lun-Wei and Martins, Andre and Srikumar, Vivek",
    booktitle = "Findings of the Association for Computational Linguistics ACL 2024",
    month = aug,
    year = "2024",
    address = "Bangkok, Thailand and virtual meeting",
    publisher = "Association for Computational Linguistics",
    url = "https://aclanthology.org/2024.findings-acl.456",
    doi = "10.18653/v1/2024.findings-acl.456",
    pages = "7655--7671",
}

@inproceedings{
swift,
title={{SWIFT}: On-the-Fly Self-Speculative Decoding for {LLM} Inference Acceleration},
author={Heming Xia and Yongqi Li and Jun Zhang and Cunxiao Du and Wenjie Li},
booktitle={The Thirteenth International Conference on Learning Representations},
year={2025},
url={https://openreview.net/forum?id=EKJhH5D5wA}
}

@misc{humaneval,
      title={Evaluating Large Language Models Trained on Code}, 
      author={Mark Chen and Jerry Tworek and Heewoo Jun and Qiming Yuan and Henrique Ponde de Oliveira Pinto and Jared Kaplan and Harri Edwards and Yuri Burda and Nicholas Joseph and Greg Brockman and Alex Ray and Raul Puri and Gretchen Krueger and Michael Petrov and Heidy Khlaaf and Girish Sastry and Pamela Mishkin and Brooke Chan and Scott Gray and Nick Ryder and Mikhail Pavlov and Alethea Power and Lukasz Kaiser and Mohammad Bavarian and Clemens Winter and Philippe Tillet and Felipe Petroski Such and Dave Cummings and Matthias Plappert and Fotios Chantzis and Elizabeth Barnes and Ariel Herbert-Voss and William Hebgen Guss and Alex Nichol and Alex Paino and Nikolas Tezak and Jie Tang and Igor Babuschkin and Suchir Balaji and Shantanu Jain and William Saunders and Christopher Hesse and Andrew N. Carr and Jan Leike and Josh Achiam and Vedant Misra and Evan Morikawa and Alec Radford and Matthew Knight and Miles Brundage and Mira Murati and Katie Mayer and Peter Welinder and Bob McGrew and Dario Amodei and Sam McCandlish and Ilya Sutskever and Wojciech Zaremba},
      year={2021},
      eprint={2107.03374},
      archivePrefix={arXiv},
      primaryClass={cs.LG},
      url={https://arxiv.org/abs/2107.03374}, 
}

@misc{gsm8k,
      title={Training Verifiers to Solve Math Word Problems}, 
      author={Karl Cobbe and Vineet Kosaraju and Mohammad Bavarian and Mark Chen and Heewoo Jun and Lukasz Kaiser and Matthias Plappert and Jerry Tworek and Jacob Hilton and Reiichiro Nakano and Christopher Hesse and John Schulman},
      year={2021},
      eprint={2110.14168},
      archivePrefix={arXiv},
      primaryClass={cs.LG},
      url={https://arxiv.org/abs/2110.14168}, 
}

@inproceedings{cnndm,
    title = "Abstractive Text Summarization using Sequence-to-sequence {RNN}s and Beyond",
    author = "Nallapati, Ramesh  and
      Zhou, Bowen  and
      dos Santos, Cicero  and
      Gu{\ensuremath{\dot{}}}l{\c{c}}ehre, {\c{C}}a{\u{g}}lar  and
      Xiang, Bing",
    editor = "Riezler, Stefan  and
      Goldberg, Yoav",
    booktitle = "Proceedings of the 20th {SIGNLL} Conference on Computational Natural Language Learning",
    month = aug,
    year = "2016",
    address = "Berlin, Germany",
    publisher = "Association for Computational Linguistics",
    url = "https://aclanthology.org/K16-1028/",
    doi = "10.18653/v1/K16-1028",
    pages = "280--290"
}

@inproceedings{transformers,
    title = "Transformers: State-of-the-Art Natural Language Processing",
    author = "Thomas Wolf and Lysandre Debut and Victor Sanh and Julien Chaumond and Clement Delangue and Anthony Moi and Pierric Cistac and Tim Rault and Rémi Louf and Morgan Funtowicz and Joe Davison and Sam Shleifer and Patrick von Platen and Clara Ma and Yacine Jernite and Julien Plu and Canwen Xu and Teven Le Scao and Sylvain Gugger and Mariama Drame and Quentin Lhoest and Alexander M. Rush",
    booktitle = "Proceedings of the 2020 Conference on Empirical Methods in Natural Language Processing: System Demonstrations",
    month = oct,
    year = "2020",
    address = "Online",
    publisher = "Association for Computational Linguistics",
    url = "https://www.aclweb.org/anthology/2020.emnlp-demos.6",
    pages = "38--45"
}

@misc{pytorch,
      title={PyTorch: An Imperative Style, High-Performance Deep Learning Library}, 
      author={Adam Paszke and Sam Gross and Francisco Massa and Adam Lerer and James Bradbury and Gregory Chanan and Trevor Killeen and Zeming Lin and Natalia Gimelshein and Luca Antiga and Alban Desmaison and Andreas Köpf and Edward Yang and Zach DeVito and Martin Raison and Alykhan Tejani and Sasank Chilamkurthy and Benoit Steiner and Lu Fang and Junjie Bai and Soumith Chintala},
      year={2019},
      eprint={1912.01703},
      archivePrefix={arXiv},
      primaryClass={cs.LG},
      url={https://arxiv.org/abs/1912.01703}, 
}

@misc{cuda,
  author={NVIDIA and Vingelmann, Péter and Fitzek, Frank H.P.},
  title={CUDA, release: 10.2.89},
  year={2020},
  url={https://developer.nvidia.com/cuda-toolkit},
}

@inproceedings{
math,
title={Measuring Mathematical Problem Solving With the {MATH} Dataset},
author={Dan Hendrycks and Collin Burns and Saurav Kadavath and Akul Arora and Steven Basart and Eric Tang and Dawn Song and Jacob Steinhardt},
booktitle={Thirty-fifth Conference on Neural Information Processing Systems Datasets and Benchmarks Track (Round 2)},
year={2021},
url={https://openreview.net/forum?id=7Bywt2mQsCe}
}

@misc{AIME,
  author       = {{Mathematical Association of America}},
  title        = {MAA Invitational Competitions},
  howpublished = {\url{https://maa.org/maa-invitational-competitions/}},
  note         = {Accessed: 2025-09-09},
  year         = {2025}
}

@inproceedings{longbench,
    title = "{L}ong{B}ench: A Bilingual, Multitask Benchmark for Long Context Understanding",
    author = "Bai, Yushi and Lv, Xin  and Zhang, Jiajie  and Lyu, Hongchang  and
      Tang, Jiankai  and Huang, Zhidian  and Du, Zhengxiao  and Liu, Xiao  and Zeng, Aohan  and Hou, Lei  and Dong, Yuxiao  and Tang, Jie  and Li, Juanzi",
    booktitle = "Proceedings of the 62nd Annual Meeting of the Association for Computational Linguistics (Volume 1: Long Papers)",
    month = aug,
    year = "2024",
    address = "Bangkok, Thailand",
    publisher = "Association for Computational Linguistics",
    url = "https://aclanthology.org/2024.acl-long.172",
    doi = "10.18653/v1/2024.acl-long.172",
    pages = "3119--3137",
}

@misc{longspec,
      title={LongSpec: Long-Context Lossless Speculative Decoding with Efficient Drafting and Verification}, 
      author={Penghui Yang and Cunxiao Du and Fengzhuo Zhang and Haonan Wang and Tianyu Pang and Chao Du and Bo An},
      year={2026},
      eprint={2502.17421},
      archivePrefix={arXiv},
      primaryClass={cs.CL},
      url={https://arxiv.org/abs/2502.17421}, 
}

@inproceedings{xiong2025long,
  title={Long-Context Modeling with Dynamic Hierarchical Sparse Attention for On-Device LLMs},
  author={Xiong, Siheng and Zou, Joe and Fekri, Faramarz and Cho, Yae Jee},
  booktitle={NeurIPS 2025 Workshop on Efficient Reasoning}
}

\appendix
\clearpage
\begin{table*}[t]
\centering
\small
\caption{Training and deployment comparison for different methods.
Example training costs: \textsc{EAGLE} $8\times$ RTX~3090 for 1--2 days; \textsc{Hydra} $8\times$ A100 for training; \textsc{Medusa} $1\times$ A100 for 5 hours.}
\setlength{\tabcolsep}{1.4mm}
\resizebox{\linewidth}{!}{
\begin{tabular}{@{}lccccc@{}}
\toprule
{\bf Methods} & {\bf Training Cost} & {\bf Additional Parameters} & {\bf Lossless Quality?} & {\bf Deployment Difficulty} \\
\midrule
\textsc{Eagle}  & High     & AR Heads            & \greenyes & High \\
\textsc{Hydra}  & High     & MLP Heads & \redno    & Moderate \\
\textsc{Medusa} & Moderate & MLP Heads                         & \redno    & Moderate \\
\rowcolor{c0!5}\method~(Ours) & \textbf{None} & \textbf{None} & \greenyes & \textbf{Plug-and-Play} \\
\bottomrule
\end{tabular}}

\label{tab:comparison}
\end{table*}

\section{More Discussions to Related Work}\label{app:related_work}
We provide additional discussion of existing works in both draft-model-based speculative decoding and draft-model-free speculative decoding. An intuitive comparison is shown in Table \ref{tab:comparison}.

\paragraph{Draft-model-based speculative decoding.} Besides the discussions in Section \ref{related_work}, we discuss additional works on draft-model-based speculative decoding. For example, HASS \citep{hass} identifies the inconsistency between EAGLE's training and inference phases, and proposes a multi-step training framework to address this. CORAL \citep{coral} proposes a cross-step representation alignment to address this problem. Judge Decoding \citep{judge} recognizes the potential of accepting high-quality but refused draft tokens to further improve the acceleration. Gumiho \citep{gumiho} demonstrates that the initial draft token is more important and proposes a hybrid model to combine serial and parallel draft heads. LongSpec \citep{longspec} proposes a long-context lossless speculative decoding framework with efficient drafting and verification.

Besides these methods that focus on the training process \citep{xiong2025long}, the process of verification also draws extensive interest, mainly focusing on adaptive draft length. SpecDec++ \citep{specdec++} formulates the verification process as a Markov decision process to adaptively determine the draft length. OPT-Tree \citep{opt-tree} proposes a method to search for the optimal tree structure that maximizes the
mathematical expectation of the acceptance length in each decoding step. AdaEAGLE \citep{adaeagle} proposes a novel framework to explicitly model the draft tree structure for EAGLE series models. PEARL \citep{pearl} pioneers this direction by serving the draft model and the target model in parallel to achieve a segmented adaptive draft length.

\paragraph{Draft-model-free speculative decoding.} For layer sparsity, SPEED \citep{speed} proposes a method to speculatively execute multiple future tokens in parallel with the current token using predicted values based on early-layer hidden states. FREE \citep{free}  proposes a shallow-deep module and a synchronized parallel decoding to improve the efficiency. EESD \citep{eesd} proposes an early-exiting framework with a self-distillation method and leverages Thompson Sampling to regulate the generation processes. For retrieval-based SD, Token Recycling \citep{tokenrecycling} proposes a method to utilize the dropped draft tokens and generate draft tokens via an adjacency matrix. Different from our method, Token Recycling is a \textbf{query-dependent} method that utilizes information from other queries, which may result in limitations when applied to real-world applications with complex and dynamic user inputs. SAM Decoding \citep{samdecoding} utilizes a common text corpus and dynamic text
 sequence as retrieved sources and proposes a suffix automaton to efficiently obtain more accurate match positions.

\section{Details of Motivation Experiments}\label{app:motivation_exp}

To investigate the prediction ability of the last logit, we conduct two motivation experiments to demonstrate its effectiveness and robustness. For the effectiveness of the last logit, as shown in Figure \ref{fig:motivation_acc}, we conduct autoregressive inference for the 6 models and record the logits for each decoded token. Then, for each decoded token $x_i$ which is sampled from the logit $p_{i-1}$, we investigate the rank of $x_i$ in $p_{i-2}$, i.e., the corresponding last logit, and visualize the statistics of the rank in Figure \ref{fig:motivation_acc}. For the robustness of the last logit, as shown in Figure \ref{fig:motivation_mat}, we conduct \textit{last logit decoding} and investigate its mean accepted tokens per decoding step. Both experiments are conducted on a small subset of Spec-Bench, where we randomly sample 2 questions for each sub-category of MT, and 10 questions for other categories (Trans, Sum, QA, Math, RAG).

\paragraph{Why Not Use Next-Next-Next Tokens (NNNT)?}
A natural question is whether we can extend the approach to predict even further tokens. We investigated this: the ground-truth NNNT appears in the top-60 last logits with $\geq$40\% probability, showing the phenomenon does extend.

However, we did \textbf{not} use NNNT due to combinatorial explosion. If we speculate top-60 for both NNT and NNNT, the draft tree would contain at least $60^2 = 3{,}600$ tokens, far exceeding efficient verification capacity. As shown in Table~\ref{tab:llama3_pruning_k} (Appendix~\ref{more_capacity}), tree capacity beyond 64-128 tokens yields diminishing returns as verification overhead dominates.

Our chosen approach (NNT + retrieval with K=64) represents the optimal efficiency-accuracy trade-off given current hardware constraints.

\paragraph{Intuition: Why Can Last Logits Predict Next-Next Tokens?}
While our contribution is an empirical discovery, we offer intuition for why this phenomenon exists:

The last logit $p_{i-1} = P(x_i | x_1, \ldots, x_{i-1})$ is computed from hidden states encoding the full context up to position $i-1$. While the model is trained to predict $x_i$, the rich contextual representation necessarily contains information about likely continuations beyond $x_i$---otherwise, the model could not generate coherent multi-token sequences.

Our findings suggest that top-ranked tokens in $p_{i-1}$ correlate with likely values of $x_{i+1}$, particularly in structured text where multi-token patterns (phrases, named entities, common collocations) are prevalent. This predictive signal exists \textit{without fine-tuning}, making it immediately exploitable for inference acceleration.

We leave formal theoretical analysis of this phenomenon to future work.

\newpage
\section{Pseudo Code to Prepare Attention Inputs}\label{app:pseudocode}

We provide pseudo code to organize the retrieved multiple draft token sequences into a draft tree and prepare its attention mask. More detailed implementations can be found in our attached code.

\begin{lstlisting}[language=Python]

def prepare_attention_inputs(past_len, next_token, candidate_list, num_draft_tokens):
    'LogitSpec organizes draft tokens in a tree manner. Each sub-sequence corresponds to a local causal mask.'

    seq_len = num_draft_tokens + 1

    # organize the candidate list into a sequence
    draft_ids = [next_token] + [token for sub in candidate_list for token in sub]

    # prepare original position ids and attention mask
    position_ids = torch.zeros((1, seq_len), dtype=torch.long)
    causal_mask = torch.full((seq_len, past_len + seq_len), fill_value=0)
    causal_mask[:, :past_len+1] = 1

    # prepare causal mask
    idx = 1
    for sub_sequence in candidate_list:
        l = len(sub_sequence)
        sub_mask = torch.tril(torch.ones((l, l)))
        causal_mask[idx:idx+l, idx+past_len:idx+past_len+l] = sub_mask
        position_ids[0, idx:idx+l] = torch.arange(l) + 1
        idx += l

    position_ids += past_len
    return draft_ids, causal_mask, position_ids

\end{lstlisting}

\section{More Implementation Details}\label{app:implementation}

\subsection{Baselines}

We compare \method against strong speculative decoding baselines. For draft-model-free, plug-and-play methods, we include: (a) Lookahead Decoding \citep{lade}, which utilizes dropped draft tokens as a retrieval corpus; (b) REST \citep{rest}, which employs an external knowledge base as the retrieval model; and (c) PLD \citep{pld}, which extracts n-grams from the current user-input prompt and decoded tokens as the retrieval model. In addition, we report results for a draft-model-based baseline, vanilla speculative sampling (SpS) \citep{sps1, sps2} with a small Vicuna-68M draft model, to illustrate the trade-off between extra model parameters / training and acceleration. {
We use Vicuna series models (7B, 13B, 33B), Llama-2 series models (7B, 13B, 70B), Llama-3.1-Instruct-8B and Qwen-3-8B as backbones, and report results with baseline methods on Vicuna and Llama-2 as the main results, where the performance of baseline methods can be reproduced with their official implementations.
}

\subsection{Hardware and Software Configurations}
We follow the setting of Spec-Bench with Pytorch \citep{pytorch} 2.6.0 and CUDA \citep{cuda} 12.4. Our implementation is based on Hugging Face transformers. For main experiments with Vicuna and Llama-2 models (Section \ref{sec:exp_set} and all baseline comparisons), we use transformers \citep{transformers} version 4.37.1. For additional experiments with Llama-3.1-8B-Instruct and Qwen3-8B (Section \ref{more_backbone}), we use transformers version 4.57.1, which is necessary to support these newer model architectures. Experiments with models between 7B and 33B are conducted on a single NVIDIA SXM A100 80G GPU, while experiments with 70B models are conducted on 2 NVIDIA SXM A100 80G GPUs. The inference precision is \texttt{float16}. Following prior work, we conduct experiments with temperature \texttt{0} and batch size 1. The maximal generation lengths are 1024.

\subsection{Retrieval Process}\label{app:retrieval}
The retrieval process significantly affects overall acceleration. A naive implementation would be computationally prohibitive for \method.

\paragraph{Naive Approach.} For each of $k$ speculated next-next tokens, scan the entire prompt (length $L$) to find matching $m$-grams, yielding $O(k \cdot m \cdot L)$ complexity.

\paragraph{Our Hash-Based Optimization.} We reduce this to $O(n + k)$ per decoding step:
\begin{enumerate}
    \item \textbf{Preprocessing (amortized):} Slide a window of size $m$ over the prompt once, building a hash table mapping each $m$-gram (key) to its continuation span (up to $n$ tokens as value). Cost: $O(L)$, done once per prompt.
    \item \textbf{Per-step retrieval:} For each of $k$ next-next token candidates:
    \begin{itemize}
        \item Construct the $m$-gram query: $O(m) \approx O(1)$ (since $m$ is small, typically 3)
        \item Hash table lookup: $O(1)$
        \item Copy continuation span: $O(n)$
    \end{itemize}
    Total per step: $O(k \cdot (1 + n)) = O(k + kn)$.
\end{enumerate}

\paragraph{Memory Cost.} The hash table requires $O(mL)$ memory, which is negligible compared to KV cache (typically several GB for long contexts).

\paragraph{Empirical Validation.} Figure~\ref{fig:runtime} shows retrieval accounts for only 1.17\% of total wall-clock time, confirming our optimization is effective in practice.

\subsection{Evaluation Instructions}\label{app:instruction}
In our experiments, we employ different instructions for different evaluation tasks. Specifically, for Spec-Bench, we use its standard instructions:

\begin{tcolorbox}[fontupper = \ttfamily, title=Prompt Templates for Spec-Bench]
    A chat between a curious user and an artificial intelligence assistant. The assistant gives helpful, detailed, and polite answers to the user's questions. \\

    USER: \mybox[rej]{Question} \\

    ASSISTANT:
\end{tcolorbox}

For CNN/DM, we prepend ``Summarize: '' to the question:

\begin{tcolorbox}[fontupper = \ttfamily, title=Prompt Templates for CNN/DM]
    A chat between a curious user and an artificial intelligence assistant. The assistant gives helpful, detailed, and polite answers to the user's questions. \\

    USER: Summarize: \mybox[rej]{Question} \\

    ASSISTANT:
\end{tcolorbox}

For GSM8K, we follow the setting with \citep{pearl} and use an 8-shot CoT for inference:

\begin{tcolorbox}[fontupper = \ttfamily, title=Prompt Templates for GSM8K]
    A chat between a curious user and an artificial intelligence assistant. The assistant gives helpful, detailed, and polite answers to the user's questions. \\

    USER: \mybox{8-shot CoT } Q: \mybox[rej]{Question} \\

    ASSISTANT:
\end{tcolorbox}

For HumanEval, we add a simple instruction ``Please help me to complete this code, just output your codes directly.'':

\begin{tcolorbox}[fontupper = \ttfamily, title=Prompt Templates for HumanEval]
    A chat between a curious user and an artificial intelligence assistant. The assistant gives helpful, detailed, and polite answers to the user's questions. \\

    USER: Please help me to complete this code, just output your codes directly. \mybox[rej]{Question} \\

    ASSISTANT:
\end{tcolorbox}

\subsection{Dataset Configurations}
In our experiments, we evaluate \method on four categories of text generation tasks, including Spec-Bench, CNN/DM, GSM8K, and HumanEval. For Spec-Bench and HumanEval, we use the full data for evaluation. For CNN/DM and GSM8K, we randomly sample 1000 questions for evaluation. The maximal generation length is set as 1024 across all the experiments.

\subsection{Model Configurations}
In our experiments, all the models are deployed with precision \texttt{float16}. The \texttt{<|eos token|>} matches the tokenizer's \texttt{<|eos token|>}.  To effectively alleviate the overhead of rolling back the KV cache to the accepted draft tokens, we follow Medusa \citep{medusa} to allocate continuous GPU memory for all the KV cache.

\section*{Ethics Statement}
This work adheres to the ACL Code of Ethics. Our study does not involve human subjects or personally identifiable information, and we only use publicly available datasets under their respective licenses. We transparently report our methods and potential risks and do not recommend deployment in high-stakes settings without further safety assessments.

\section{More In-depth Analysis of \method}\label{app:more_exp_results}

\subsection{Comparison with Draft-Model-Based Methods}\label{app:draft_model_based}
For completeness, we also compare \method against the draft-model-based methods EAGLE~\citep{eagle} and Medusa~\citep{medusa} on Spec-Bench with Vicuna-7B. As shown in Table~\ref{tab:draft_model_based}, \method attains a speedup comparable to EAGLE and clearly above Medusa, while being entirely training-free and requiring no extra parameters.
\begin{table}[!h]
\centering
\caption{Comparison with draft-model-based methods on Spec-Bench with \textbf{Vicuna-7B}. \method is training-free, while EAGLE and Medusa require training extra heads.}
\label{tab:draft_model_based}
\resizebox{\linewidth}{!}{
\begin{tabular}{lccl}
    \toprule
    Method & MAT & Speedup & Note \\
    \midrule
    EAGLE    & 3.58 & $2.10\times$ & Requires training \& extra heads \\
    Medusa   & 2.31 & $1.71\times$ & Requires training \& extra heads \\
    \blue{\textbf{\method}} & \blue{\textbf{2.44}} & \blue{\textbf{$2.01\times$}} & \textbf{Training-free \& plug-and-play} \\
    \bottomrule
\end{tabular}}
\end{table}

\subsection{More Results on Different LLM Backbones} \label{more_backbone}

\paragraph{Comparison with State-of-the-Art Retrieval-Based Methods.}
To position \method against recent advances in retrieval-based speculative decoding, we compare with SAM Decoding \citep{samdecoding} and Token Recycling \citep{tokenrecycling}. Both methods are plug-and-play and do not require draft models, making them directly comparable to \method.

Table~\ref{tab:advanced_baselines_llama3} shows results on Llama-3.1-8B-Instruct across AIME, LongBench, and MATH. \method achieves 7.5\% higher average speedup (2.58$\times$ vs 2.40$\times$ for Token Recycling, 2.34$\times$ for SAM). The key advantage comes from \method's next-next token speculation:

\begin{itemize}
    \item \textbf{vs. Token Recycling}: Token Recycling reuses previously rejected draft tokens via an adjacency matrix, which is query-dependent and requires cross-query information. In contrast, \method is query-independent and uses the target model's own predictive capability to guide retrieval, making it more robust when query patterns are diverse.
    \item \textbf{vs. SAM Decoding}: SAM uses suffix automata for efficient n-gram matching from a static corpus. While this improves matching speed, it still faces the fundamental challenge of low match rates when context repetition is limited. \method addresses this by speculating next-next tokens, effectively expanding the retrieval space beyond observed n-grams.
\end{itemize}

Notably, on AIME (a challenging reasoning dataset with low repetition), \method achieves 3.09$\times$ speedup compared to 2.82$\times$ for Token Recycling, demonstrating the value of predictive guidance when retrieval alone struggles.

Here, we provide additional results for different LLM backbones to provide more insights into \method, including Llama 2 chat series models, LLaMA-3.1-8B-Instruct, and Qwen3-8B.

We present the results of Llama 2 chat series models on Spec-Bench, CNN/DM, GSM8K, and HumanEval in Table \ref{tab:llama2_3_bench} and Table \ref{tab:llama2_specbench}. \method consistently achieves the best acceleration among all baselines across Spec-Bench, CNN/DM, GSM8K, and HumanEval, further validating its effectiveness. For completeness, Table~\ref{tab:vicuna_specbench} reports the per-subtask Spec-Bench breakdown on Vicuna 7B/13B/33B, complementing the summary statistics in the main text.

\subsection{Per-Task Spec-Bench Breakdown on Vicuna}\label{app:vicuna_specbench}
\paragraph{Per-task Analysis.} The detailed breakdown reveals important insights into \method's robustness across diverse tasks. In the Translation task, where almost no reference is available to serve as draft tokens, PLD and Lookahead achieve poor acceleration. Even equipped with a large external database ($\approx$ 12 GB), REST only achieves $1.17\times\sim 1.31\times$ speedup, while \method achieves $1.38\times\sim 1.43\times$ speedup without external databases, demonstrating the prediction ability of the last logit on the next next token. In contrast, on tasks with high context repetition such as Summarization and RAG, \method achieves even stronger performance, highlighting how next-next token speculation complements retrieval-based approaches across varying task characteristics.

\begin{table*}[!t]
\centering
\caption{Experimental results of \method on Spec-Bench \citep{spec-bench} with \textbf{Vicuna}. We report the speedup ratio on each sub task, mean accepted tokens per decoding step (MAT) and overall speedup ratio. We \textbf{bold} the best results and \underline{underline} the suboptimal results for each backbone model. \\}
\label{tab:vicuna_specbench}
\resizebox{\linewidth}{!}{
\begin{tabular}{lccccccc|cc}
\toprule
\textbf{Models}   & \textbf{Method}      & \textbf{MT}    & \textbf{Trans} &\textbf{Sum} & \textbf{QA} & \textbf{Math} & \textbf{RAG} & \textbf{MAT}  & \textbf{Speedup}\\
\midrule
\multirow{5}{*}{\textbf{Vicuna 7B}}

    & Lookahead  & 1.40 & 1.14 & 1.19 & 1.24 & 1.55 & 1.09 & 1.66 & 1.27 \\
    & REST   & 1.63 & \underline{1.31} & 1.36 & \underline{1.66} & 1.21 & \underline{1.73} & 1.82 & 1.48 \\
    & PLD  & 1.64 & 1.04 & \underline{2.43} & 1.14 & \underline{1.61} & 1.71 & 1.73 & \underline{1.59} \\
    & SpS & \underline{1.66} & 1.13 & 1.62 & 1.49 & 1.47 & 1.55 & \underline{2.28} & 1.49 \\
    & \blue{\textbf{\method}}   & \blue{\textbf{1.92}} & \blue{\textbf{1.39}} & \blue{\textbf{2.68}} & \blue{\textbf{1.70}} & \blue{\textbf{2.22}} & \blue{\textbf{1.87}} & \blue{\textbf{2.44}} & \blue{\textbf{2.01}} \\
    \midrule
\multirow{5}{*}{\textbf{Vicuna 13B}}
    & Lookahead  & 1.30 & 1.06 & 1.20 & 1.12 & 1.48 & 1.12 & 1.63 & 1.22 \\
    & REST   & 1.52 & \underline{1.17} & 1.37 & \textbf{1.53} & 1.19 & 1.55 & 1.82 & 1.38 \\
    & PLD  & 1.47 & 1.02 & \underline{2.19} & 1.03 & \underline{1.57} & \underline{1.71} & 1.68 & 1.48 \\
    & SpS & \underline{1.60} & 1.13 & 1.68 & 1.39 & 1.53 & 1.67 & \underline{2.18} & \underline{1.49} \\
    & \blue{\textbf{\method}}   & \blue{\textbf{1.89}} & \blue{\textbf{1.43}} & \blue{\textbf{2.33}} & \blue{\underline{1.43}} & \blue{\textbf{2.23}} & \blue{\textbf{1.93}} & \blue{\textbf{2.32}} & \blue{\textbf{1.93}} \\
\midrule
\multirow{5}{*}{\textbf{Vicuna 33B}}
    & Lookahead  & 1.32 & 1.08 & 1.20 & 1.06 & 1.54 & 1.15 & 1.61 & 1.24 \\
    & REST   & 1.63 & 1.27 & 1.45 & \textbf{1.61} & 1.30 & 1.61 & 1.80 & 1.48 \\
    & PLD  & 1.44 & 1.06 & \underline{2.00} & 1.07 & 1.55 & 1.45 & 1.55 & 1.42 \\
    & SpS & \underline{1.75} & \underline{1.28} & 1.76 & \underline{1.53} & \underline{1.69} & \underline{1.68} & \underline{2.01} & \underline{1.61} \\
& \blue{\textbf{\method}}   & \blue{\textbf{1.77}} & \blue{\textbf{1.38}} & \blue{\textbf{2.15}} & \blue{1.37} & \blue{\textbf{2.00}} & \blue{\textbf{1.69}} & \blue{\textbf{2.13}} & \blue{\textbf{1.76}} \\

\bottomrule
\end{tabular}
}
\end{table*}

Table~\ref{tab:advanced_baselines_llama3} compares \method against SAM Decoding and Token Recycling on AIME, LongBench, and MATH with the Llama-3.1-8B-Instruct backbone under identical decoding settings ($T=0$, $K=64$, greedy sampling). \method consistently attains the best MAT and speedup across all datasets.

\begin{table*}[!h]
\centering
\caption{Comparison with stronger retrieval-based baselines on \textbf{Llama-3.1-8B-Instruct}. We report MAT and end-to-end speedup on AIME, LongBench, and MATH, as well as the macro average. All methods share the same hardware, decoding parameters (temperature $T=0$), and tree capacity $K=64$.}
\label{tab:advanced_baselines_llama3}
\resizebox{\linewidth}{!}{
\begin{tabular}{lcccccccccc}
    \toprule
    \multirow{2}{*}{Method} & \multicolumn{2}{c}{\textbf{AIME}} & \multicolumn{2}{c}{\textbf{LongBench}} & \multicolumn{2}{c}{\textbf{MATH}} & \multicolumn{2}{c}{\textbf{Average}} \\
    \cmidrule(lr){2-3} \cmidrule(lr){4-5} \cmidrule(lr){6-7} \cmidrule(lr){8-9}
     & MAT & Speedup & MAT & Speedup & MAT & Speedup & MAT & Speedup \\
    \midrule
    AR (Greedy) & 1.00 & 1.00$\times$ & 1.00 & 1.00$\times$ & 1.00 & 1.00$\times$ & 1.00 & 1.00$\times$ \\
    SAM Decoding & 3.11 & 2.73$\times$ & 3.01 & 2.04$\times$ & 2.66 & 2.26$\times$ & 2.92 & 2.34$\times$ \\
    Token Recycling & 3.10 & 2.82$\times$ & 2.96 & 1.97$\times$ & 2.80 & 2.40$\times$ & 2.96 & 2.40$\times$ \\
    \blue{\textbf{\method}} & \blue{\textbf{3.57}} & \blue{\textbf{3.09$\times$}} & \blue{\textbf{3.08}} & \blue{\textbf{2.10$\times$}} & \blue{\textbf{3.05}} & \blue{\textbf{2.56$\times$}} & \blue{\textbf{3.23}} & \blue{\textbf{2.58$\times$}} \\
    \bottomrule
\end{tabular}}
\end{table*}

We also present more results applying LLaMA-3.1-8B-Instruct and Qwen3-8B on Spec-Bench, CNN/DM, GSM8K, and HumanEval in Tables \ref{tab:llama3_qwen3_specbench} and \ref{tab:llama3_qwen3_bench}. We observe that compared to standard autoregressive decoding, \method consistently achieves around $2\times$ inference acceleration on average across these datasets.

\begin{table*}[!htbp]
\centering
\caption{Experimental results of \method on CNN/DM \citep{cnndm}, GSM8K \citep{gsm8k} and HumanEval \citep{humaneval} with \textbf{Llama-2}. We report the mean accepted tokens per decoding step (MAT) and overall speedup ratio. We \textbf{bold} the best results and \underline{underline} the suboptimal results for each backbone model. \\}
\label{tab:llama2_3_bench}
\resizebox{\linewidth}{!}{
\begin{tabular}{lccccccc|cc}
\toprule
\multirow{2}{*}{\textbf{Models}} & \multirow{2}{*}{\textbf{Method}} & \multicolumn{2}{c}{\textbf{CNN/DM}} & \multicolumn{2}{c}{\textbf{GSM8K}} & \multicolumn{2}{c}{\textbf{HumanEval}} & \multicolumn{2}{c}{\textbf{Overall}} \\
\cmidrule(lr){3-4} \cmidrule(lr){5-6} \cmidrule(lr){7-8} \cmidrule(lr){9-10}
& & MAT  & Speedup & MAT  & Speedup & MAT  & Speedup & MAT  & Speedup\\

\midrule
\multirow{5}{*}{\textbf{Llama 2 7B}}
    & Lookahead  & 1.58 & 1.36 & 2.02 & 1.67 & 1.77 & \underline{1.53} & 1.79 & 1.52 \\
    & REST & 1.71 & 1.33 & 1.51 & 1.18 & 1.97 & 1.52 & 1.73 & 1.34\\
    & PLD  & 1.89 & \underline{1.73} & \underline{3.32} & \underline{2.98} & 1.57 & 1.41 & 2.26 & \underline{2.04}\\
    & SpS  & \underline{1.99} & 1.46 & 2.83 & 1.87 & \underline{2.13} & 1.51 & \underline{2.32} & 1.61\\
    & \blue{\textbf{\method}} & \blue{\textbf{2.41}} & \blue{\textbf{2.02}} & \blue{\textbf{4.44}} & \blue{\textbf{3.68}} & \blue{\textbf{2.17}} & \blue{\textbf{1.75}} & \blue{\textbf{3.01}} & \blue{\textbf{2.48}}\\

\midrule
\multirow{5}{*}{\textbf{Llama 2 13B}}
    & Lookahead  & 1.56 & 1.18 & 2.08 & 1.52 & 1.84 & 1.66 & 1.83 & 1.45 \\
    & REST & 1.71 & 1.34 & 1.53 & 1.26 & 1.96 & 1.63 & 1.73 & 1.41\\
    & PLD  & 1.89 & \underline{1.52} & \underline{3.24} & \underline{2.54} & 1.73 & 1.63 & 2.29 & \underline{1.90}\\
    & SpS  & \underline{1.95} & 1.34 & 2.87 & 1.85 & \underline{2.33} & \underline{1.76} & \underline{2.38} & 1.65\\
    & \blue{\textbf{\method}} & \blue{\textbf{2.43}} & \blue{\textbf{2.03}} & \blue{\textbf{4.31}} & \blue{\textbf{3.24}} & \blue{\textbf{2.38}} & \blue{\textbf{2.10}} & \blue{\textbf{3.04}} & \blue{\textbf{2.46}} \\

\midrule
\multirow{5}{*}{\textbf{Llama 2 70B}}
    & Lookahead  & 1.53 & 1.28 & 1.90 & 1.57 & 1.86 & 1.57 & 1.76 & 1.47\\
    & REST & 1.67 & 1.35 & 1.63 & 1.32 & 1.96 & 1.66 & 1.75 & 1.44\\
    & PLD  & 1.98 & \underline{1.74} & 1.63 & 1.46 & 1.62 & 1.49 & 1.74 & 1.56\\
    & SpS  & \underline{2.01} & 1.71 & \underline{1.98} & \underline{1.69} & \underline{2.21} & \underline{1.70} & \underline{2.07} & \underline{1.70}\\
    & \blue{\textbf{\method}} & \blue{\textbf{2.67}} & \blue{\textbf{2.10}} & \blue{\textbf{2.37}} & \blue{\textbf{1.87}} & \blue{\textbf{2.33}} & \blue{\textbf{1.93}} & \blue{\textbf{2.46}} & \blue{\textbf{1.97}}\\

\bottomrule
\end{tabular}
}
\end{table*}

\begin{table*}[!h]
\centering
\caption{Experimental results of \method on Spec-Bench \citep{spec-bench} with \textbf{Llama-2}. We report the speedup ratio on each sub task, mean accepted tokens per decoding step (MAT) and overall speedup ratio. We \textbf{bold} the best results and \underline{underline} the suboptimal results for each backbone model. \\}
\label{tab:llama2_specbench}
\resizebox{\linewidth}{!}{
\begin{tabular}{lccccccc|cc}
\toprule
\textbf{Models}   & \textbf{Method}      & \textbf{MT}    & \textbf{Trans} &\textbf{Sum} & \textbf{QA} & \textbf{Math} & \textbf{RAG} & \textbf{MAT}  & \textbf{Speedup}\\
\midrule
\multirow{5}{*}{\textbf{Llama 2 7B}}
    & Lookahead  & 1.55 & \underline{1.44} & 1.42 & 1.48 & \underline{1.69} & 1.46 & 1.69 & 1.51 \\
    & REST   & 1.58 & 1.20 & 1.38 & \underline{1.61} & 1.31 & 1.55 & 1.83 & 1.48 \\
    & PLD  & 1.46 & 1.34 & \underline{1.89} & 1.23 & 1.65 & 1.58 & 1.59 & 1.49 \\
    & SpS & \underline{1.57} & 1.38 & 1.55 & 1.46 & 1.55 & \underline{1.60} & \underline{2.07} & \underline{1.53} \\
    & \blue{\textbf{\method}}   & \blue{\textbf{1.83}} & \blue{\textbf{1.72}} & \blue{\textbf{2.19}} & \blue{\textbf{1.63}} & \blue{\textbf{2.15}} & \blue{\textbf{1.94}} & \blue{\textbf{2.15}} & \blue{\textbf{1.87}} \\

\midrule
\multirow{5}{*}{\textbf{Llama 2 13B}}
    & Lookahead  & 1.39 & \underline{1.36} & 1.21 & 1.39 & \underline{1.69} & 1.25 & 1.68 & 1.37 \\
    & REST   & \underline{1.53} & 1.14 & 1.30 & \underline{1.51} & 1.23 & 1.45 & 1.85 & 1.42 \\
    & PLD  & 1.36 & 1.19 & \underline{1.58} & 1.16 & 1.62 & 1.37 & 1.56 & 1.37 \\
    & SpS & 1.52 & 1.26 & 1.43 & 1.42 & 1.54 & \underline{1.50} & \underline{2.03} & \underline{1.48} \\
    & \blue{\textbf{\method}}   & \blue{\textbf{1.73}} & \blue{\textbf{1.57}} & \blue{\textbf{1.86}} & \blue{\textbf{1.53}} & \blue{\textbf{2.14}} & \blue{\textbf{1.73}} & \blue{\textbf{2.12}} & \blue{\textbf{1.74}} \\

\midrule
\multirow{5}{*}{\textbf{Llama 2 70B}}
    & Lookahead  & 1.45&1.35&1.28&1.38&\underline{1.71}&1.31&1.66&1.41 \\
    & REST   & 1.63&1.33&1.38&\textbf{1.67}&1.35&1.55&1.83&1.53 \\
    & PLD  & 1.34&1.32&\underline{1.76}&1.18&1.63&1.47&1.51&1.39 \\
    & SpS & \underline{1.65}&\underline{1.50}&1.62&1.57&1.70&\underline{1.68}&\underline{1.88}&\underline{1.63}\\
    & \blue{\textbf{\method}}   & \blue{\textbf{1.66}} & \blue{\textbf{1.58}} & \blue{\textbf{1.95}} & \blue{\underline{1.58}} & \blue{\textbf{2.03}} & \blue{\textbf{1.78}} & \blue{\textbf{2.12}} & \blue{\textbf{1.72}} \\
\bottomrule
\end{tabular}
}
\end{table*}

\begin{table*}[!h]
\centering
\caption{Experimental results of \method on Spec-Bench \citep{spec-bench} with \textbf{ LLaMA-3.1-8B-Instruct and Qwen3-8B}. We report the speedup ratio on each sub task, mean accepted tokens per decoding step (MAT) and overall speedup ratio. We \textbf{bold} the best results for each backbone model. \\}
\resizebox{\linewidth}{!}{
\begin{tabular}{llcccccc|cc}
    \toprule
    \textbf{Model} & \textbf{Method} & \textbf{MT} & \textbf{Trans} & \textbf{Sum} & \textbf{QA} & \textbf{MATH} & \textbf{RAG} & \textbf{MAT} & \textbf{Speedup} \\
    \midrule
    Llama-3.1-8B-Instruct & Vanilla & 1.00 & 1.00 & 1.00 & 1.00 & 1.00 & 1.00 & 1.00 & 1.00 \\
    {{Llama-3.1-8B-Instruct}} & \blue{\textbf{\method}} & \blue{\textbf{1.89}} & \blue{\textbf{1.67}} & \blue{\textbf{1.94}} &\blue{\textbf{ 1.68}} & \blue{\textbf{2.01 }}& \blue{\textbf{1.77}} &\blue{\textbf{2.11}} & \blue{\textbf{1.88}} \\
    \midrule
    Qwen-3-8B & Vanilla & 1.00 & 1.00 & 1.00 & 1.00 & 1.00 & 1.00 & 1.00 & 1.00 \\
    {{Qwen-3-8B}} & \blue{\textbf{\method}} & \blue{\textbf{1.71}} & \blue{\textbf{1.74}} & \blue{\textbf{1.64}} & \blue{\textbf{1.65}} & \blue{\textbf{1.89}} & \blue{\textbf{1.68}} & \blue{\textbf{1.95}} & \blue{\textbf{1.75}} \\
    \bottomrule
\end{tabular}
  \label{tab:llama3_qwen3_specbench}
}
\end{table*}

\begin{table*}[!h]
\centering
\caption{Experimental results of \method on CNN/DM \citep{cnndm}, GSM8K \citep{gsm8k} and HumanEval \citep{humaneval} with \textbf{LLaMA-3.1-8B-Instruct and Qwen3-8B}. We report the mean accepted tokens per decoding step (MAT) and overall speedup ratio. We \textbf{bold} the best results for each backbone model. \\}
\label{tab:llama3_qwen3_bench}
\resizebox{\linewidth}{!}{
\begin{tabular}{@{}llcccccccc@{}}
    \toprule
    \multirow{2}{*}{Model} & \multirow{2}{*}{Method} & \multicolumn{2}{c}{\textbf{CNN/DM}} & \multicolumn{2}{c}{\textbf{GSM8K}} & \multicolumn{2}{c}{\textbf{Humaneval}} & \multicolumn{2}{c}{\textbf{Overall}} \\
    \cmidrule(lr){3-4} \cmidrule(lr){5-6} \cmidrule(lr){7-8} \cmidrule(lr){9-10}
     & & MAT & Speedup & MAT & Speedup & MAT & Speedup & {MAT} & {Speedup} \\
    \midrule
    Llama-3.1-8B-Instruct & Vanilla  & 1.00 & 1.00 & 1.00 & 1.00 & 1.00 & 1.00 & {1.00} & {1.00} \\
    Llama-3.1-8B-Instruct & \blue{\textbf{\method}}  & \blue{\textbf{2.04}} & \blue{\textbf{1.85}} & \blue{\textbf{2.18}} & \blue{\textbf{1.95}} & \blue{\textbf{2.63}} & \blue{\textbf{2.31}} & \blue{\textbf{{2.28}}} & \blue{\textbf{{2.04}}} \\

    \midrule
    Qwen-3-8B              & Vanilla  & 1.00 & 1.00 & 1.00 & 1.00 & 1.00 & 1.00 & {1.00} & {1.00} \\
    Qwen-3-8B              & \blue{\textbf{\method}}  & \blue{\textbf{1.77}} & \blue{\textbf{1.59}} & \blue{\textbf{2.18}} & \blue{\textbf{1.94}} & \blue{\textbf{2.18}} & \blue{\textbf{1.92}} & \blue{\textbf{{2.04}}}& \blue{\textbf{{1.82}}} \\

    \bottomrule
\end{tabular}
}
\end{table*}

\subsection{More Results on Different Benchmarks}\label{more_benchmark}
\begin{table}[t]
  \centering
  \caption{Results of \method on AIME datasets.}
  \label{tab:llama3_AIME}
  \begin{tabular}{@{}lcccc@{}}
    \toprule
    \multirow{2}{*}{Method} & \multicolumn{2}{c}{AIME24} & \multicolumn{2}{c}{AIME25} \\
    \cmidrule(lr){2-3} \cmidrule(lr){4-5}
     & MAT & Speedup & MAT & Speedup \\
    \midrule
    Vanilla   & 1.00 & 1.00 & 1.00 & 1.00 \\
    \method & 3.41 & 3.25 & 3.76 & 3.33 \\
    \bottomrule
  \end{tabular}
\end{table}

\begin{figure*}[t]
    \centering
    \includegraphics[width=0.9\textwidth]{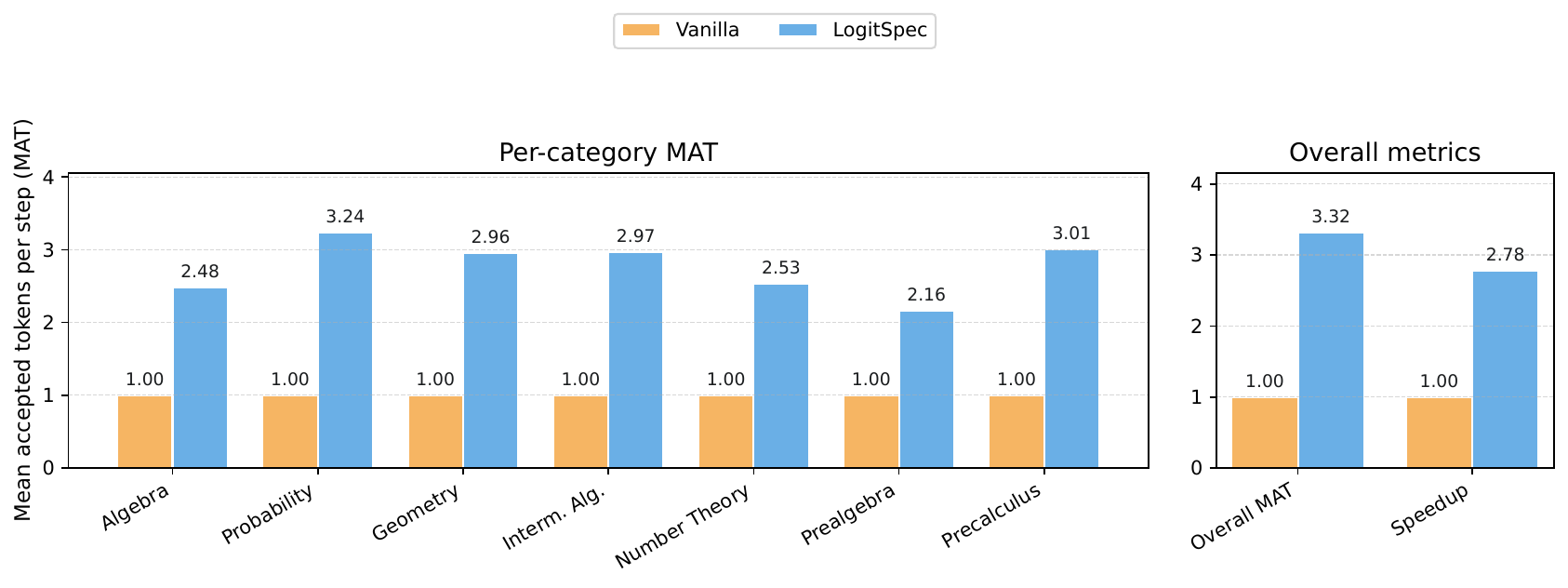}
    \caption{LogitSpec performance on MATH benchmark subsets.}
    \label{fig:math_bars}
\end{figure*}

We provide more results for \method on complex math reasoning tasks including MATH \citep{math} and AIME 24 \& 25 \citep{AIME} datasets in Figure \ref{fig:math_bars} and Table \ref{tab:llama3_AIME}. For each of these benchmarks, we report the real-world speedup for each subset, the overall mean accepted tokens per decoding step (MAT), and the overall speedup. These expanded experimental results further corroborate \method's effectiveness and applicability, yielding an overall MAT of 3.32 and an overall speedup of $2.78\times$ on MATH, and up to 3.76 MAT and $3.33\times$ speedup on AIME25. This demonstrates \method's robust performance even on challenging reasoning tasks.

\subsection{Sampling with Non-Zero Temperatures}\label{app:sampling}

\paragraph{Why LogitSpec Remains Effective Under Sampling.}
A key question is whether LogitSpec's reliance on top-$k$ logits makes it sensitive to temperature scaling. We clarify the mechanism and provide empirical validation.

\textbf{Candidate Selection is Temperature-Invariant.} In our implementation, speculative next-next token candidates are selected from the \textbf{raw logits} before applying temperature. Since temperature scaling is a monotonic transformation, it preserves ranking:
\begin{align}
z_i > z_j \quad &\Longleftrightarrow \quad \exp(z_i/T) > \exp(z_j/T) \nonumber\\
&\Longleftrightarrow \quad p_i(T) > p_j(T), \quad \forall T > 0
\end{align}
where $p(T) = \text{softmax}(z/T)$. Therefore, the top-$k$ candidate set used for speculation is \textbf{identical} across all temperatures and does not ``fill with noise tokens'' as $T$ increases.

\textbf{What Changes: Acceptance Probability.} Higher temperatures make the actual sampled trajectory deviate more from the most-likely path, reducing acceptance rates. This is an inherent property of all speculative decoding methods under sampling, not specific to LogitSpec.

To study the robustness of \method under stochastic decoding, we evaluate performance across temperatures $T\in\{0.2,0.4,0.6,0.8\}$ on AIME, LongBench, and MATH with Llama-3.1-8B-Instruct. As shown in Table~\ref{tab:sampling_ablation}, \method maintains strong performance even at higher temperatures, still achieving speedups above $1.8\times$ even at $T=0.8$. This demonstrates that \method remains effective across diverse generation scenarios.

\begin{table*}[!h]
\centering
\caption{LogitSpec under different sampling temperatures on Llama-3.1-8B-Instruct. Draft candidates are drawn from raw logits before temperature scaling, so the speculative candidate set is unchanged across $T$; higher $T$ only changes the sampled path.}
\label{tab:sampling_ablation}
\resizebox{0.95\linewidth}{!}{
\begin{tabular}{lcccccc}
    \toprule
    Temperature & AIME MAT & AIME Speedup & LongBench MAT & LongBench Speedup & MATH MAT & MATH Speedup \\
    \midrule
    0.0 & 3.57 & 3.09$\times$ & 3.08 & 2.10$\times$ & 3.05 & 2.56$\times$ \\
    0.2 & 3.37 & 2.85$\times$ & 2.94 & 1.94$\times$ & 3.20 & 2.64$\times$ \\
    0.4 & 3.51 & 2.96$\times$ & 2.92 & 1.93$\times$ & 3.03 & 2.55$\times$ \\
    0.6 & 3.26 & 2.76$\times$ & 2.87 & 1.89$\times$ & 2.91 & 2.46$\times$ \\
    0.8 & 3.15 & 2.66$\times$ & 2.84 & 1.86$\times$ & 2.81 & 2.42$\times$ \\
    \bottomrule
\end{tabular}}
\end{table*}

\subsection{Cross-Lingual Next-Next Token Coverage}\label{app:multilingual}
To validate the universality of the last logit's predictive capability across different linguistic structures, we conduct experiments on the MMMLU benchmark covering five languages: English, Chinese, Japanese, German, and French. Table~\ref{tab:multilingual_hit_rate} reports the fraction of ground-truth next-next tokens appearing in the top-$k$ logits for each language. The results show consistent coverage across all languages: at $k=64$, over $50\%$ of next-next tokens fall within the top-$k$ predictions for every language. This demonstrates that our observation about the last logit's predictive capability is not restricted to English or specific linguistic structures, but rather represents a fundamental property of modern LLMs that generalizes across diverse languages with different word orders and morphological characteristics.

\begin{table}[!h]
\centering
\caption{Fraction of ground-truth next-next tokens that appear within the top-$k$ logits of the last token on MMMLU subsets (Llama-3.1-8B-Instruct). Each language averages 20 questions.}
\label{tab:multilingual_hit_rate}
\resizebox{0.9\linewidth}{!}{
\begin{tabular}{lccccc}
    \toprule
    Language & $k=4$ & $k=8$ & $k=16$ & $k=32$ & $k=64$ \\
    \midrule
    English & 15.25\% & 22.02\% & 33.60\% & 42.06\% & 51.83\% \\
    Chinese & 17.14\% & 24.21\% & 36.29\% & 45.68\% & 55.07\% \\
    Japanese & 15.16\% & 24.30\% & 34.72\% & 46.17\% & 55.87\% \\
    German & 14.64\% & 21.44\% & 32.87\% & 42.42\% & 50.64\% \\
    French & 14.86\% & 21.77\% & 32.21\% & 42.78\% & 52.22\% \\
    \bottomrule
\end{tabular}}
\end{table}

These results show that the last logit's predictive capability is not restricted to English or specific linguistic structures. The consistency across languages with markedly different characteristics---Chinese (logographic, subject-verb-object), Japanese (agglutinative, subject-object-verb), German (inflectional, flexible word order), and French (Romance, subject-verb-object)---suggests this is a fundamental property of modern LLMs' internal representations rather than a language-specific artifact. Combined with our diverse task coverage (reasoning, QA, summarization, translation, code generation), this cross-lingual validation provides strong evidence for the generality of our core observation.

\subsection{More Results on Long-Context Scenarios}\label{more_longdata}
\begin{figure*}[t]
    \centering
    \includegraphics[width=0.9\textwidth]{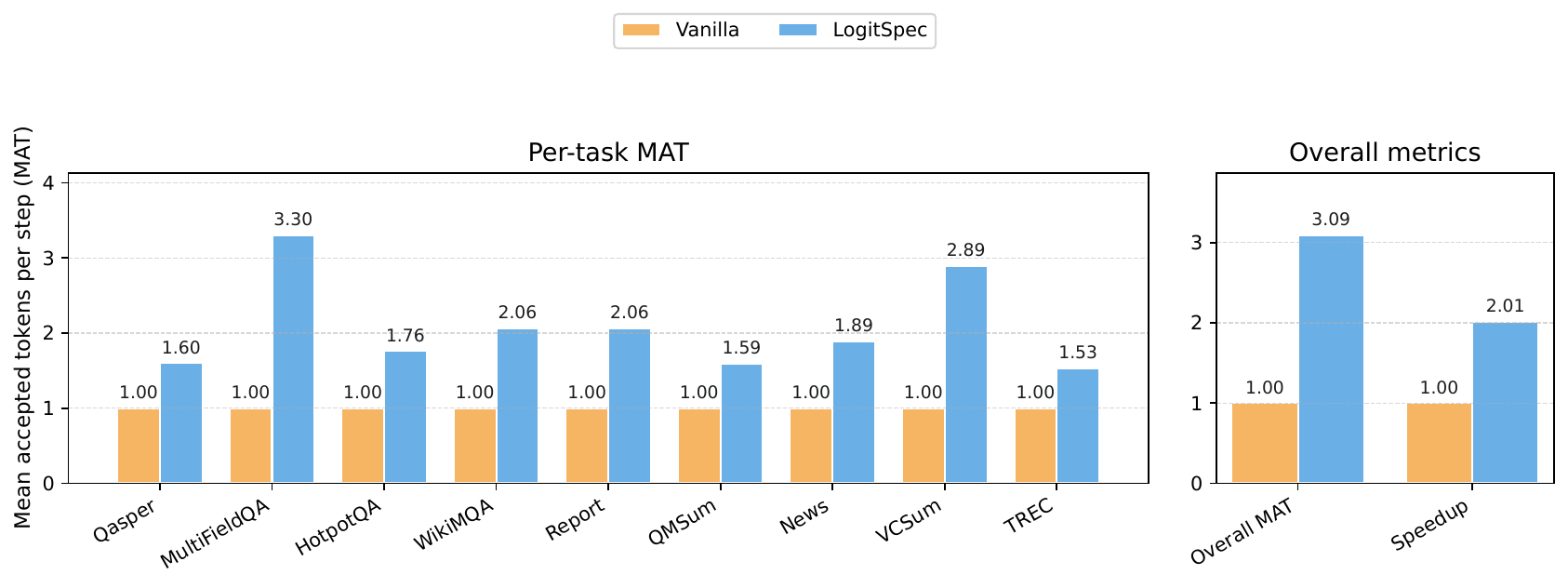}
    \caption{LogitSpec performance on LongBench subsets.}
    \label{fig:longbench_bars}
\end{figure*}

We conduct a new set of experiments on LongBench using Llama-3.1-8B-Instruct as the backbone model, randomly sampling 100 problems for evaluation. We observe that \method is highly effective in these long-context scenarios. As shown in Figure \ref{fig:longbench_bars}, our method achieves an overall MAT of $3.09$ and an overall speedup of $2.01$.
This strong performance is directly linked to the core mechanics of our method. \method's retrieval model is built from the user-input prompt and previously decoded tokens. A longer context generally provides a richer database for this retrieval process. However, a large context also increases the chance of finding ambiguous or incorrect n-grams, which is precisely where \method's ``next next token speculation'' offers a significant advantage. By using a more specific multi-token query, it effectively disambiguates the retrieval process, which is particularly crucial in a long context with many repetitive phrases. Furthermore, our retrieval implementation was designed for efficiency, using a hash table to ensure that the overhead remains negligible (around 1.17$\%$ of the total decoding time) even as the sequence length grows. Therefore, these new results confirm that \method is a robust and effective solution for accelerating inference in demanding long-context scenarios.

\subsection{Ablation Study on Different Pruning Strategies}\label{more_pruning}
In Section \ref{draft_sec}, we apply a heuristic pruning strategy to control the number of draft tokens. The role of the pruning algorithm is to effectively control the size of the resulting draft tree, keeping the decoding overhead low without sacrificing too many possibilities, which is a strategy to balance breadth and cost. To provide more insight into the pruning strategy, we conduct ablation studies in Table \ref{tab:llama3_pruning_ab}. Specifically, we consider two different pruning strategies:

\begin{itemize}
  \item [(a)] Strategy $1$: A rank-based heuristic: if the rank is $<$$4$, we preserve $5$ tokens; $<$$8$, $4$ tokens; $<$$16$, $3$ tokens; $<$32, $2$ tokens; else $1$ token.
  \item [(b)] Strategy $2$: A simple heuristic: preserving $4$ tokens for all speculated next next tokens.
\end{itemize}

The results in Table \ref{tab:llama3_pruning_ab} demonstrate that while different pruning strategies have some effect on the final performance, the overall performance of \method is quite robust. The speedup remains stable at approximately $1.9\times$ across these different approaches.

We also would like to clarify that the core contribution of our work is \textbf{next next token speculation} using the last logit to guide and enhance the efficiency and accuracy of retrieval-based speculative decoding. The primary purpose of the pruning algorithm is to serve as an auxiliary module for our core mechanism.
\begin{table*}[!h]
\centering
\caption{Ablation study on pruning strategies. \method$_{with \hspace{1mm} s1}$ and \method$_{with \hspace{1mm} s2}$ denote \method with Strategy 1 and Strategy 2, respectively. \\}
\label{tab:llama3_pruning_ab}
\resizebox{\linewidth}{!}{
   \begin{tabular}{@{}lcccccc|cc@{}}
    \toprule
    \textbf{Method}  & \textbf{MT} & \textbf{Trans} & \textbf{Sum} & \textbf{QA} & \textbf{MATH} & \textbf{RAG} & \textbf{MAT} & \textbf{Speedup} \\
    \midrule

    \method$_{with \hspace{1mm} s1}$                                       & \textbf{1.90} & \textbf{1.68} & \textbf{1.94} & 1.66 & \textbf{2.04} & \textbf{1.78} & \textbf{2.12} & \textbf{1.88} \\
    \method$_{with \hspace{1mm} s2}$                                       & 1.86 & 1.67 & 1.87 & \textbf{1.68} & 2.03 & 1.74 & 2.02 & 1.85 \\
    \blue{\textbf{\method} \textbf{(ori)}}     & \blue{{1.89}} & \blue{{1.67}} & \blue{\textbf{1.94}} & \blue{\textbf{1.68}} & \blue{{2.01}} & \blue{{1.77}} & \blue{{2.11}} & \blue{\textbf{1.88}} \\
    \bottomrule
  \end{tabular}
}
\end{table*}

\subsection{Ablation Study on Draft Tree Capacity}\label{more_capacity}
As mentioned in Section \ref{draft_sec}, we set the capacity of the draft tree to K=64. To provide further insight, we evaluate a range of $K$ values from $32$ to $128$ in Table \ref{tab:llama3_pruning_k}. These results reveal a clear trade-off: as K increases, the MAT per step improves (from $2.03$ to $2.30$), since a larger tree offers more opportunities for token acceptance. However, the overall speedup ratio peaks at K=64 ($1.92\times$) and subsequently declines. This is because verifying an overly large draft tree introduces significant computational overhead that ultimately negates the advantages of a higher acceptance rate.

\begin{table*}[!h]
\centering
\caption{Ablation study on pruning hyperparameters $K$ using LLaMA-3.1-8B-Instruct as the backbone model. \\}
\label{tab:llama3_pruning_k}
\resizebox{\linewidth}{!}{
  \begin{tabular}{@{}lcccccc|cc@{}}
    \toprule
    \textbf{Method} & \textbf{MT} & \textbf{Trans} & \textbf{Sum} & \textbf{QA} & \textbf{MATH} & \textbf{RAG} & \textbf{MAT} & \textbf{Speedup} \\
    \midrule
    Vanilla & 1.00 & 1.00 & 1.00 & 1.00 & 1.00 & 1.00 & 1.00 & 1.00 \\
    \method$_{32}$ & 1.79 & 1.62 & 1.82 & 1.56 & 1.93 & 1.67 & 1.95 & 1.79 \\
    \method$_{48}$ & 1.86 & 1.66 & 1.90 & 1.67 & 2.00 & 1.73 & 2.04 & 1.85 \\
    \blue{\textbf{\method$_{64}$}} & \blue{\textbf{1.89}} & \blue{\textbf{1.67}} & \blue{\textbf{1.94}} & \blue{\textbf{1.68}} & \blue{\textbf{2.01}} & \blue{\textbf{1.77}} & \blue{2.11} & \blue{\textbf{1.88}} \\
     \method$_{80}$ & 1.85 & 1.64 & 1.86 & 1.67 & 1.84 & 1.74 & 2.14 & 1.83 \\
    \method$_{96}$ & 1.81 & 1.65 & 1.83 & 1.66 & 1.84 & 1.73 & 2.17 & 1.80 \\
    \method$_{112}$ & 1.79 & 1.61 & 1.77 & 1.64 & 1.83 & 1.69 & 2.21 & 1.77 \\
    \method$_{128}$ & 1.74 & 1.62 & 1.73 & 1.58 & 1.81 & 1.62 & \textbf{2.22} & 1.73 \\

    \bottomrule
  \end{tabular}
}
\end{table*}

\subsection{Real-World Example}\label{more_realworld}
To provide more insights into \method, we present a real-world example to illustrate how ``next next token speculation'' allows \method to succeed where standard retrieval methods fail.

Taking the prefix

\begin{lstlisting}
Q: A pen costs as much as a pencil and eraser combined. A pencil costs $1.20 and an eraser costs $0.30. How much will 8 pens cost?
A: To find the cost of 8 pens, we first need to find the cost of one pen. A pencil costs $1.20 and an eraser costs $0.30. The
\end{lstlisting}

At the first step, \method generates the next token ``combined'' and then speculates the next next token with the following high-probability candidates:
\begin{itemize}
    \item .
    \item total
    \item cost
    \item combination
    \item pen
    \item eraser
\end{itemize}

Then, \method extends each candidate with retrieved n-grams and verifies these draft sequences:

\begin{itemize}
    \item \textbf{[\(\times\)]} . A pencil costs
    \item \textbf{[\(\times\)]} total they had
    \item \textbf{[\(\checkmark\)]} cost of one pen
    \item \textbf{[\(\times\)]} combination (no matched n-grams retrieved)
    \item \textbf{[\(\times\)]} pen.
    \item \textbf{[\(\times\)]} eraser costs \$
\end{itemize}

Finally, with the guidance of the last logit, we successfully accept $3$ draft tokens and generate ``the combined cost of one pen''. In the next decoding step, \method again accepts $3$ draft tokens and generates ``the combined cost of one pencil and one eraser''. However, without last-logit speculation, standard retrieval can only retrieve the first n-gram ``A pencil costs''.

\section{Discussion: Integration with Production Frameworks}\label{app:vllm}

\paragraph{vLLM Integration Feasibility.}
While our current implementation uses Hugging Face transformers, several reviewers asked about integration with production inference frameworks like vLLM. We discuss technical feasibility:

\textbf{Tree Attention Support.} vLLM has recently added native tree attention support:
\begin{itemize}
    \item Flash attention kernel for tree structures: \url{https://github.com/vllm-project/flash-attention/pull/81}
    \item Official tree attention backend: \url{https://docs.vllm.ai/en/latest/api/vllm/v1/attention/backends/tree_attn/}
\end{itemize}

\textbf{Compatibility with vLLM's Architecture.} LogitSpec's draft tree can be naturally represented as multiple sequences with shared prefixes, aligning with vLLM's batched inference paradigm. Each branch corresponds to a different draft sequence, similar to the prefill process for multiple sequences. The nano-PEARL repository demonstrates a similar integration approach for tree-based verification in vLLM.

\textbf{Minimal Overhead.} Beyond tree attention, LogitSpec only requires:
\begin{itemize}
    \item The last logit at each step (already computed by vLLM)
    \item CPU-based retrieval and tree construction (can run in parallel with GPU inference)
\end{itemize}

We conclude that there are no fundamental algorithmic obstacles to deploying LogitSpec in vLLM; the remaining work is engineering integration, which we leave to future work.

\section{LLM Usage}
We used a large language model (LLM)--based writing assistant solely for grammar and wording improvements on draft text. The LLM did not generate research ideas, claims, proofs, algorithms, code, figures, or analyses, and it did not have access to any non-public data. All edits suggested by the LLM were manually reviewed and either accepted or rewritten by the authors, who take full responsibility for the final content. The LLM is not an author of this paper.

\end{document}